\documentclass[sigconf]{acmart}
\acmSubmissionID{711}
\usepackage{bm}

\AtBeginDocument{%
  \providecommand\BibTeX{{%
    \normalfont B\kern-0.5em{\scshape i\kern-0.25em b}\kern-0.8em\TeX}}}

\copyrightyear{2023} 
\acmYear{2023} 
\setcopyright{rightsretained} 
\acmConference[SIGGRAPH '23 Conference Proceedings]{Special Interest Group
on Computer Graphics and Interactive Techniques Conference Conference
Proceedings}{August 6--10, 2023}{Los Angeles, CA, USA}
\acmBooktitle{Special Interest Group on Computer Graphics and Interactive
Techniques Conference Conference Proceedings (SIGGRAPH '23 Conference
Proceedings), August 6--10, 2023, Los Angeles, CA, USA}
\acmDOI{10.1145/3588432.3591545}
\acmISBN{979-8-4007-0159-7/23/08}

\begin{document}

\title{LatentAvatar: Learning Latent Expression Code for Expressive Neural Head Avatar}

\author{Yuelang Xu}
\affiliation{%
  \institution{Tsinghua University}
  \city{Beijing}
  \country{China}}
\email{xll20@mails.tsinghua.edu.cn}
\orcid{0009-0001-6834-8199}

\author{Hongwen Zhang}
\affiliation{%
  \institution{Tsinghua University}
  \city{Beijing}
  \country{China}}
\email{zhanghongwen@mail.tsinghua.edu.cn}
\orcid{0000-0001-8633-4551}

\author{Lizhen Wang}
\affiliation{%
  \institution{Tsinghua University}
  \institution{NNKosmos Technology}
  \city{Beijing}
  \country{China}}
\email{wlz18@mails.tsinghua.edu.cn}
\orcid{0000-0002-6674-9327}

\author{Xiaochen Zhao}
\affiliation{%
  \institution{Tsinghua University}
  \institution{NNKosmos Technology}
  \city{Beijing}
  \country{China}}
\email{zhaoxc19@mails.tsinghua.edu.cn}
\orcid{0000-0001-8976-7723}

\author{Han Huang}
\affiliation{%
  \institution{OPPO Research Institute}
  \city{Beijing}
  \country{China}}
\email{huang.h92@outlook.com}
\orcid{0000-0002-9278-2382}

\author{Guojun Qi}
\affiliation{%
  \institution{OPPO Research Seattle}
  \institution{Westlake University}
  \city{Seattle}
  \country{United States of America}}
\email{guojunq@gmail.com}
\orcid{0000-0003-3508-1851}

\author{Yebin Liu}
\affiliation{%
  \institution{Tsinghua University}
  \city{Beijing}
  \country{China}}
\email{liuyebin@mail.tsinghua.edu.cn}
\orcid{0000-0003-3215-0225}

\renewcommand{\shortauthors}{Yuelang Xu, et al.}

\begin{abstract}
Existing approaches to animatable NeRF-based head avatars are either built upon face templates or use the expression coefficients of templates as the driving signal.
Despite the promising progress, their performances are heavily bound by the expression power and the tracking accuracy of the templates.
In this work, we present LatentAvatar, an expressive neural head avatar driven by latent expression codes.
Such latent expression codes are learned in an end-to-end and self-supervised manner without templates, enabling our method to get rid of expression and tracking issues.
To achieve this, we leverage a latent head NeRF to learn the person-specific latent expression codes from a monocular portrait video, and further design a Y-shaped network to learn the shared latent expression codes of different subjects for cross-identity reenactment. 
By optimizing the photometric reconstruction objectives in NeRF, the latent expression codes are learned to be 3D-aware while faithfully capturing the high-frequency detailed expressions.
Moreover, by learning a mapping between the latent expression code learned in shared and person-specific settings, LatentAvatar is able to perform expressive reenactment between different subjects.
Experimental results show that our LatentAvatar is able to capture challenging expressions and the subtle movement of teeth and even eyeballs, which outperforms previous state-of-the-art solutions in both quantitative and qualitative comparisons. Project page: https://www.liuyebin.com/latentavatar.

\end{abstract}



\begin{teaserfigure}
  \includegraphics[width=\textwidth]{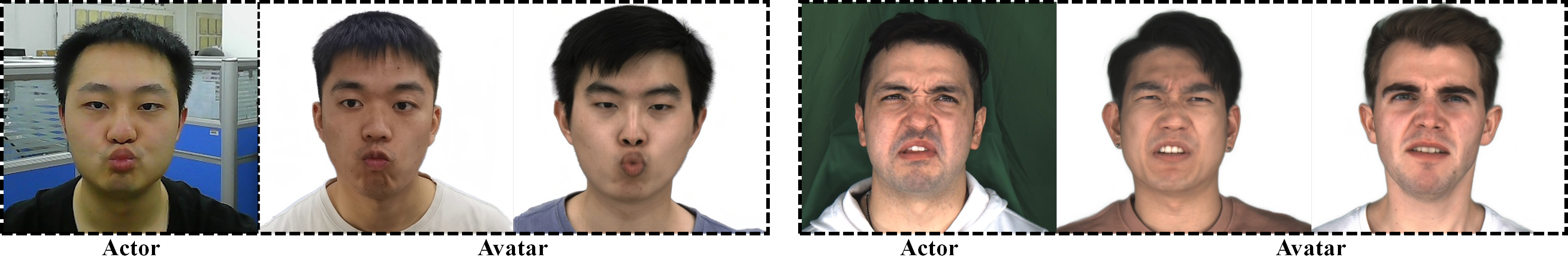}
  \vspace{-8mm}
  \caption{We propose LatentAvatar, an expressive neural head avatar driven by latent expression codes. LatentAvatar is able to capture subtle expressions such as pouting (left) and perform expressive reenactment (right) between different subjects.}
  \label{fig:teaser}
\end{teaserfigure}

\maketitle

\section{Introduction}
Creating a 3D head avatar from a monocular video has a great application prospect in digital human, CG Filmmaking, VR and AR, etc. 
This field has attracted growing attention in recent years. By leveraging the face template prior~\cite{gerig2018morphable, li2017learning} or the implicit field representation~\cite{park2019deepsdf, mildenhall2020nerf}, recent works~\cite{gafni2021dynamic, zheng2022imavatar, gao2022reconstructing, grassal2022neural, xu2023avatarmav} can recover photo-realistic 3D head avatars using a monocular video. However, despite the promising progress, efficient and expressive control of the head avatar remains unsolved in previous approaches.

When modeling a head avatar, existing methods typically leverage explicit mesh templates~\cite{zheng2022imavatar, grassal2022neural, Khakhulin2022realistic} or neural implicit representations~\cite{gafni2021dynamic, gao2022reconstructing, xu2023avatarmav, athar2021flame}.
Despite the efficiency of face templates, the expression representation power is bound by the linear expression  blendshapes or the linear skinning of the face model, which leads to coarse control of avatars and the lack of person-specific detailed expressions.
On the other hand, neural implicit representations bypass the constraint of explicit templates and synthesize the facial images directly.
However, modeling dynamic 3D heads remain challenging for implicit representations.
To control the head avatar, these methods typically resort to conditioning the implicit field with additional expression information, such as the expression coefficients of 3DMM~\cite{gerig2018morphable}, FLAME~\cite{li2017learning}, or FaceWarehouse~\cite{cao2014facewarehouse}.
Although the avatars created by these methods are not confined by the template topology, the linear expression coefficients of the face templates make it difficult to model high-frequency and detailed person-specific expressions.
Moreover, when using the face templates, the misalignment of the tracking results introduces additional deviation in the expression condition.
Meanwhile, the expression coefficients and identity coefficients of the face template are easy to be coupled with each other, leading to unexpected artifacts during the cross-identity reenactment of avatars.


To overcome these limitations, we propose LatentAvatar, an expressive neural head avatar driven by latent expression codes.
The core idea of LatentAvatar is to learn a latent expression code as the drive signal to animate the head avatar expressively.
To achieve this, we first learn a latent head NeRF, where a person-specific latent expression code is learned to drive a customized head NeRF using a monocular portrait video.
In the latent head NeRF, the expression code is learned in a self-supervised manner by the photometric reconstruction loss.
Such a latent expression code can faithfully capture those high-frequency detailed expressions of the target subject.
Besides, by driving the head radiance field, the expression code is also 3D-aware and enables the modeling of viewpoint-consistent avatars.
Compared with previous solutions, our method gets rid of the template tracking and expression issues as the latent expression code is learned in a full end-to-end manner without templates.


For cross-identity reenactment, LatentAvatar further leverages a shared latent expression code for the modeling of the shared expression between different subjects.
To this end, we introduce a Y-shaped network architecture~\cite{yan2018video, naruniec2020high} consisting one single shared encoder and two individual decoders.
The shared encoder takes both the avatar and actor images as input to learn a shared latent expression code, which is decoded by two decoders for the individual reconstruction of the input subjects.
Finally, a mapping MLP is used to build a bridge between two latent spaces by mapping the shared expression codes to the person-specific one.
In this way, LatentAvatar leverages the latent expression code learned in the shared and person-specific settings and enables expressive reenactment between different subjects.

In summary, the contributions of this work can be listed as:
\begin{itemize}
    \item We propose LatentAvatar, an expressive neural head avatar driven by latent expression codes. Such an expression code is learned in an end-to-end and self-supervised manner without templates, enabling our method to get rid of expression and tracking issues.
    \item  We leverage a latent head NeRF to learn the person-specific latent expression code from a monocular portrait video. The latent expression code is learned to drive the head NeRF, making it 3D-aware while faithfully capturing the high-frequency detailed expressions.
    \item We further leverage a Y-shaped network to learn a shared latent expression code of different subjects to enable cross-identity reenactment. By bridging the latent expression code learned in shared and person-specific settings, LatentAvatar is able to perform expressive reenactment between different subjects and surpass the performances of previous 2D and 3D solutions.

    
    
\end{itemize}

\section{Related Works}

\subsection{Head Avatar Modeling}
Reconstructing 3D head avatars from monocular videos is an appealing yet challenging task. In the past years, mainstream methods~\cite{cao2015real, cao2016real, ichim2015dynamic, hu2017avatar, deng2019accurate, nagano2018pagan} reconstruct mesh-based head avatars based on the morphable face templates tracked in the training portrait video. 
To handle highly non-rigid contents such as hair, gazes, and teeth, recent methods~\cite{grassal2022neural, Khakhulin2022realistic} leverage neural networks to learn head avatars with dynamic texture and geometry upon the FLAME mesh model~\cite{li2017learning}. 
However, these methods often result in blurred textures due to geometric inaccuracy.
To leverage more flexible representations such as the neural implicit fields~\cite{park2019deepsdf, occupancy2019mescheder}, IMavatar~\cite{zheng2022imavatar} proposes to learn head avatars with implicit geometry and texture model~\cite{li2017learning}, thus gets rid of the topology limitation of the mesh templates.
IMavatar is further extended in PointAvatar~\cite{zheng2023pointavatar} to combine the explicit point cloud with the implicit representation to improve the quality of the rendered images.
Since the emergence of NeRF representations~\cite{mildenhall2020nerf,park2021nerfies,park2021hypernerf}, there are several attempts to exploit its rendering power for neural head modeling~\cite{guo2021ad, liu2022semantic, gafni2021dynamic, wang2022morf, athar2021flame, athar2022rignerf, zheng2023avatarrex}.
Recently, FDNeRF~\cite{zhang2022fdnerf} then combine the few-shot NeRF methods~\cite{yu2020pixelnerf, chen2021mvsnerf} to extend these NeRF-based head avatar methods to few-shot reconstruction.
For faster avatar modeling, recent methods~\cite{gao2022reconstructing, xu2023avatarmav, zielonka2022instant} introduce voxel representation~\cite{mueller2022instant, fang2022fast} to solve the problem of low training speed of the NeRF model. 
However, most of the above solutions rely on the tracked faces templates as they typically use the expression coefficients of templates as the drive signal for avatar animation. 
Recent methods~\cite{hong2022headnerf, zhuang2022mofanerf, yenamandra2020i3dmm, sun2023next3d, sun2021fenerf, chan2022efficient, sun2022ide} use large-scale face datasets to train an implicit head template.
But it is not easy for these methods to achieve reenactment between different subjects.

There are also research efforts devoted to creating high-fidelity mesh-based head avatars from single-person multi-view synchronized video data~\cite{lombardi2018deep,lombardi2019neural, bi2021deep, wang2021learning, ma2021pixel}.
To capture expression information, encoder-decoder networks are typically leveraged to learn latent codes as the compact representations of inputs~\cite{lombardi2019neural,chu2020expressive,cao2021real}.
Note that these latent codes are person-specific and only used as the self-driven signal for avatar control.
To achieve high-resolution rendering, MVP representation~\cite{lombardi2021mixture} constructs a mixture of volumetric primitives on a coarse mesh and synthesizes high-fidelity images through volumetric rendering in a similar way to NeRF~\cite{mildenhall2020nerf}. 
Based on the MVP representation, Cao et al.~\cite{cao2022authentic} train a generalized head model on a super large-scale multi-view multi-person video dataset. 
Despite the high-quality results, these methods heavily rely on large-scale multi-view data and even require depth cameras for accurate tracking of the mesh model.

\begin{figure*}[ht]
  \includegraphics[width=\linewidth]{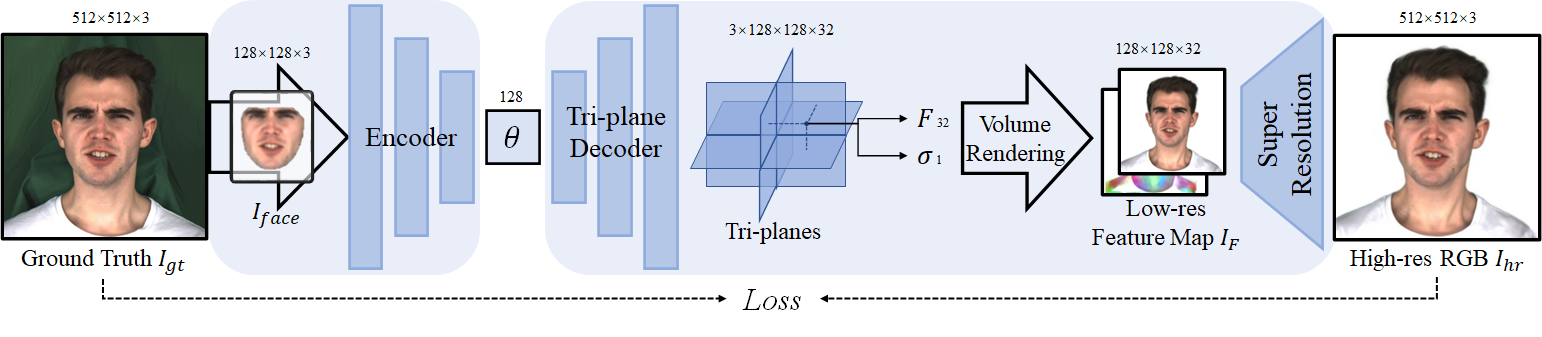}
  \vspace{-3mm}
  \caption{Overview of the Latent Head NeRF. 
  Given a portrait video, we first encode the face image to the latent expression code $\theta$, which is used as a condition to generate the tri-plane features. Given a 3D position, the feature vector $H$ is extracted from the tri-plane features for the volume rendering of the low-resolution image and feature map. finally, a super-resolution network is used to generate the corresponding high-resolution images.}
  \label{fig:autoencoder}
  \vspace{-3mm}
\end{figure*}

\subsection{Facial Reenactment}

Existing facial reenactment methods can be roughly divided into three categories: template-based, warping-based, and mapping-based methods. 
Early works~\cite{weise2011realtime, vlasic2005face, thies2015real, cao2014displaced, cao20133d, thies2016face2face, li2012a} are typically template-based~\cite{dale2011video, nirkin2018on, olszewski2017realistic} and require a source video for training, which make full use of face priors to fit the target subject with a common morphable template. 
Recent template-based works~\cite{koujan2020head2head, doukas2021head2head++, kim2018deep,zakharov2019few, wang2023styleavatar} use image-to-image generation networks~\cite{goodfellow2014generative} to synthesize photo-realistic images with the template guidance~\cite{gerig2018morphable}. These methods are extended to few-shot input setting, and further the guidance of template are replaced by semantic maps and facial landmarks~\cite{chen2020puppeteergan, zakharov2019few, nirkin2019fsgan, natsume2019fsnet, natsume2018rsgan, korshunova2017fast, yucheng2019joint, ivan2020deepfacelab}
Warping-based methods~\cite{avebuch2017bringing, siarohin2019first, geng2018warp, yin2022styleheat, wiles2018x2face} are typically few-shot, such that they require a source image or several frames as input. Give the source image, these methods do not reconstruct the face geometry but directly estimate the 2D warping from the source image to the target image.
However, learning accurate facial warping across different head poses is very challenging.
To alleviate this issue, face templates are also introduced and served as the guidance of the warping~\cite{doukas2021headgan, ren2021pirenderer}.
Recently, 3D flow fields or volumetric features are also leveraged in~\cite{drobyshev2022megaportraits, wang2021one} to generate avatar images under different head poses.
On the contrary, mapping-based methods~\cite{yan2018video, moser2021semi, naruniec2020high} directly learn the mapping between the face images of different subjects without using templates or warping maps. 
Despite the promising results achieved by the above image-based face reenactment methods, they require large-scale datasets for training and typically suffer from the lack of view consistency.

\section{Latent Head NeRF}
\label{sec:avatar}


LatentAvatar consists of a latent head NeRF, which can be driven by a latent expression code.
The latent expression code is learned in an end-to-end and self-supervised manner along with the head NeRF, for the goal of capturing person-specific detailed expressions.

\subsection{Formulation}
Previous work~\cite{gafni2021dynamic, gao2022reconstructing, guo2021ad, liu2022semantic} has exploited the power of NeRF~\cite{mildenhall2020nerf} for 3D head modeling.
A typical head NeRF model $\Phi$ can be formulated as an expression-conditioned implicit field:
\begin{equation}
 (c, \sigma) = \Phi(x, d, \theta),
\end{equation}
where $x$ and $d$ are the query point and view direction used in volumetric rendering, $c$ and $\sigma$ are the color and the density respectively, and $\theta$ denotes the expression condition.
To control the NeRF-based head model, previous methods~\cite{gafni2021dynamic, gao2022reconstructing} typically use the expression coefficients of face templates such as 3DMM~\cite{gerig2018morphable} as the expression condition $\theta$.
In these methods, the expression coefficients are obtained by tracking face templates from the input images and are not learnable during the NeRF optimization.
This process introduces two intractable limitations: i) the insufficient expression ability of 3DMM model itself makes it difficult to cover high-frequency details of person-specific expressions, and ii) the inaccuracy of the face tracking methods incurs additional deviations in the expression condition.

To tackle these issues, we propose a latent head NeRF, which does not use any pretrained face templates to model the expressions of head avatars but learns the expression condition in a latent space jointly with the NeRF optimization.
Previous work~\cite{hong2022headnerf} has tried naively learning the expression latent space in NeRF without constraints, but inversion is needed to retrieve the latent code needs and introduces additional difficulty in the avatar reenactment.
In contrast, our method learns the latent code from face images and treats the latent head NeRF as an autoencoder.
Specifically, the face image is firstly encoded into the latent expression space and then decoded via volume rendering of the radiance field. Formally, the latent head NeRF can be formulated as:
\begin{equation}
 \theta_{p} = E(I_{face}),~~ \text{and}
\end{equation}
\begin{equation}
 (c, \sigma) = \Phi(x, d, \theta_{p}),
\end{equation}
where $E$ denotes the encoder, $I_{face}$ denotes the face region of the portrait image, and $\theta_{p}$ denotes the person-specific latent expression code.
Once the autoencoder is trained, we can simply feed face images of the avatar into the encoder and use the resulting latent expression code to drive the head NeRF avatar.

Compared with the previous head NeRF~\cite{gafni2021dynamic, gao2022reconstructing}, the expression condition of our head NeRF is a learnable latent code instead of the pre-defined expression coefficients of face templates.
With the photometric reconstruction objectives, the latent expression code is learned to capture the finer-grained details of person-specific expressions during the NeRF optimization.
In our experiments, we will show that our latent expression code can even capture the subtle movement of teeth, tongue, and eyeballs of the subject, which goes far beyond the PCA-based expression coefficients of the face templates such as 3DMM~\cite{gerig2018morphable}.


\subsection{NeRF-based Decoder}
In this part, we present the detailed architecture of our NeRF-based decoder $\Phi$. 
In the generic dynamic NeRF~\cite{gafni2021dynamic}, the rendered images tend to be blurred due to the insufficient ability of pixel-wise feature learning.
To tackle this, we explore a tri-plane representation and hybrid rendering~\cite{chan2022efficient} in our latent head NeRF. 
Specifically, we feed the latent expression code $\theta_p$ to a StyleGAN-based 2D convolutional network~\cite{karras2021a} to generate tri-plane features $(\mathcal{H}_{xy}, \mathcal{H}_{yz}, \mathcal{H}_{xz})$.
Given a 3D position $x$, three feature vectors $(H_{xy}, H_{yz}, H_{xz})$ are queried by projecting it onto each of the three feature planes. 
Then, an aggregated feature vector $H$ is obtained by summing up the three vectors. 
The feature vector $H$ is further processed by a lightweight MLP for the generation of the color $c$, the density $\sigma$, and a high dimensional color feature $F$. 
Given a camera pose, we first render a low-resolution face image $I_{lr}$, a feature map $I_{F}$ through the volume rendering based on the color $c$ and feature $F$, respectively.
Meanwhile, a mask map $M$ with the same resolution is also generated during the rendering.
Then, the low-resolution face image $I_{lr}$ is first concatenated with the feature map $I_{F}$ and fed into a super-resolution module to generate a high-resolution RGB portrait image $I_{hr}$. 
In contrast to the purely generative tasks in EG3D~\cite{chan2022efficient}, our head avatar pays more attention to the rich expression enhancement of a specific person. 
Thus, the model size of the tri-plane generator can be largely reduced.
In our solution, a U-net structure with downsampling layers is used in the super-resolution module.

\subsection{Latent Code Optimization}
For end-to-end learning of the latent expression code, we jointly optimize the parameters of the encoder and the decoder of the latent head NeRF. The total loss is :
\begin{equation}
\begin{split}
\mathcal{L} = ||I_{hr} - I_{gt}||_{1} + \lambda_{vgg} VGG(I_{hr}, I_{gt}) \\
+ \lambda_{lr} ||I_{lr} - I_{gt}||_{1} + \lambda_{mask} ||M - M_{gt}||_{2},
\end{split}
\end{equation}
where $I_{gt}$ and $M_{gt}$ denote the preprocessed ground-truth image and mask respectively, $I_{lr}$ denotes the first three channels of the low-resolution feature map $I_{F}$, $M$ denotes the rendered low-resolution mask, $I_{hr}$ denotes the final high-resolution image, and $\lambda$ denotes the weight of each term. 
Besides, a VGG perceptual loss~\cite{zhang2018the} $VGG(\cdot)$ is also used during the training.


\section{Cross-Identity Reenactment}
\label{sec:cross-identity reenactment}
\begin{figure*}[ht]
  \includegraphics[width=\linewidth]{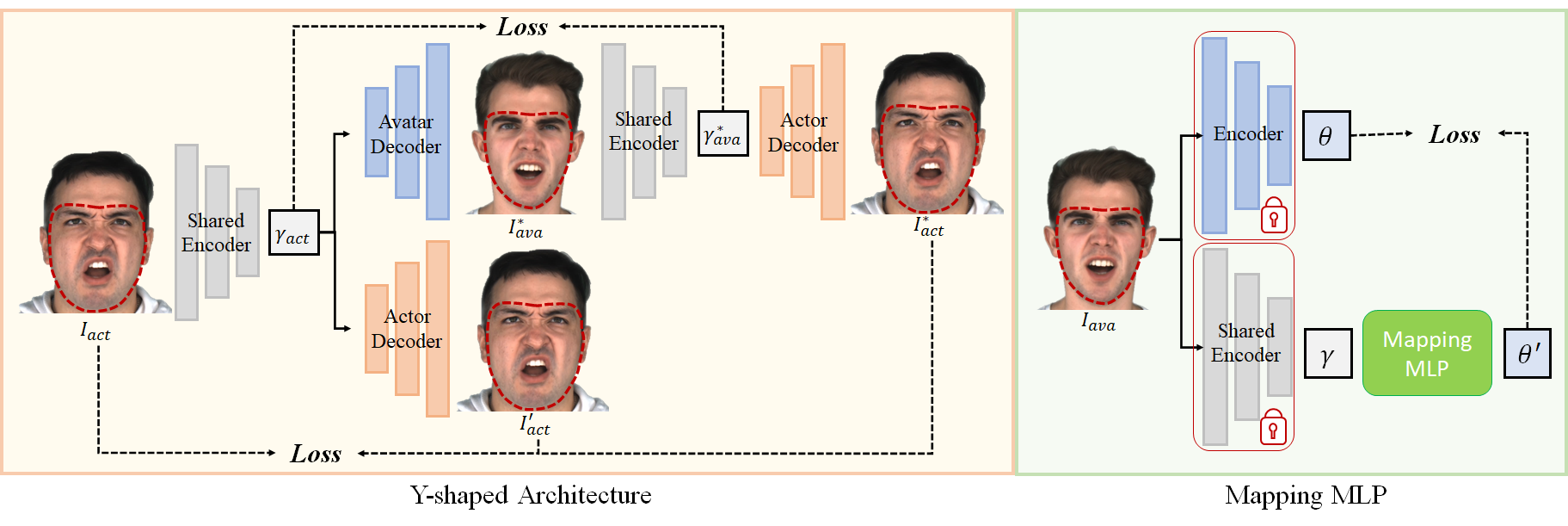}
  \vspace{-3mm}
  \caption{Illustration of the Y-shape network (left) used to learn the shared latent expression code and the mapping MLP (right) used to map the shared latent code to its person-specific one.
  In the Y-shape network, the shared encoder $E_{shared}$ encodes the input face images as the shared expression latent code, which will be decoded by the avatar and actor decoders ${D_{ava}, D_{act}}$ to the face images of the avatar and the actor individually.
  To bridge the shared and person-specific latent space, the mapping MLP learns to map the shared latent expression code to the person-specific one. 
  }
  \label{fig:y-shaped_mapping}
  \vspace{-3mm}
\end{figure*}

\begin{figure}[t]
  \includegraphics[width=\linewidth]{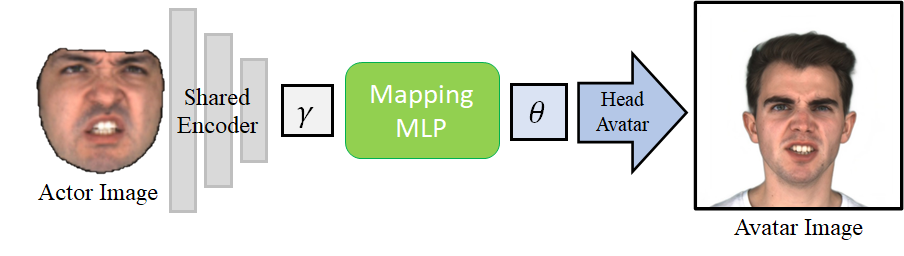}
  \vspace{-3mm}
  \caption{The process of the cross-identity reenactment in our method. The face image of the actor is first fed into the shared encoder to obtain the shared latent code $\gamma$, which is mapped as the person-specific latent code $\theta$ to drive the NeRF-based head avatar.}
  \label{fig:reenactment}
  \vspace{-3mm}
\end{figure}

Existing NeRF-based head avatar methods~\cite{gafni2021dynamic, xu2023avatarmav, gao2022reconstructing} use the expression coefficients of templates as the drive signal, which can be easily used to achieve cross-identity reenactment.
In our LatentAvatar, the drive signal is the latent expression code learned in a person-specific setting. 
The person-specific latent expression code of different subjects can not be exchanged for reenactment because they are learned separately.
To enable cross-identity reenactment, we further leverage a Y-shaped network to learn a shared latent expression code and then map it to the person-specific one as the drive signal.

\begin{figure*}[ht]
  \centering
  \includegraphics[width=1.0\linewidth]{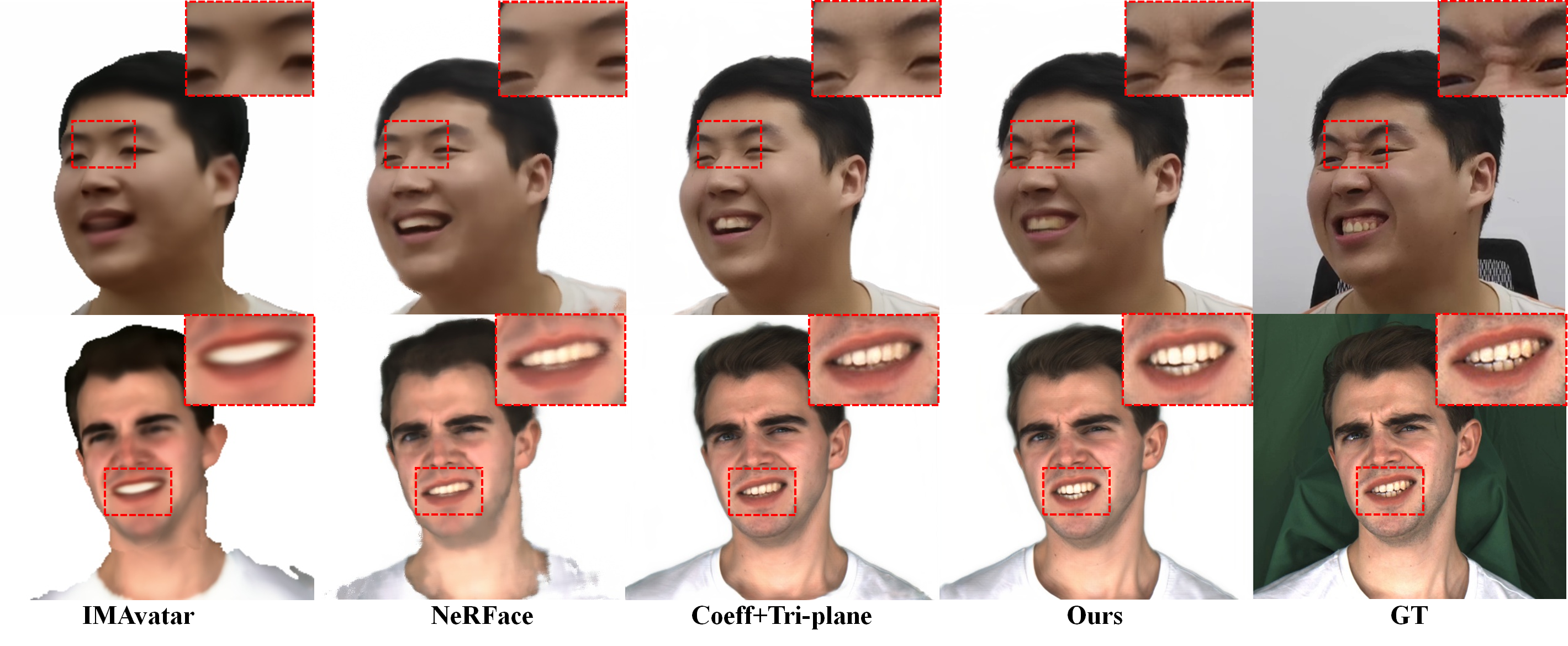}
  \vspace{-3mm}
  \caption{Qualitative comparisons of different methods on the self reenactment task. From left to right: IMavatar~\cite{zheng2022imavatar}, NeRFace~\cite{gafni2021dynamic}, Coeff+Tri-plane, and Ours. Our method surpasses other methods in the ability to capture and reproduce detailed expressions such as the wrinkles around the nose and the exposure level of teeth.}
  \vspace{-1mm}
  \label{fig:self reenactment}
\end{figure*}

\subsection{Y-shaped Network}
\label{sec:y-shaped-net}


Given monocular videos of the avatar and the actor, the shared latent expression code should allow face swapping of these two subjects while ensuring expression consistency. 
Inspired by previous work~\cite{yan2018video,naruniec2020high}, we construct a network with a Y-shaped architecture, which contains a shared encoder $E_{shared}$ and two separate decoders ${D_{ava}, D_{act}}$.
As shown in Fig.~\ref{fig:y-shaped_mapping}, the shared encoder $E_{shared}$ encodes the face images of both subjects into the \textbf{shared expression latent space} $\gamma$. Then, the avatar and actor decoders ${D_{ava}, D_{act}}$ map the latent code to the face images of the two subjects individually.


The Y-shaped network is learned with the reconstruction task of both the actor and avatar images.
For more efficient learning, we only optimize the parameters of the shared encoder $E_{shared}$ and the selected decoder of the avatar or actor, while freezing the parameters of another decoder. 
More specifically, in each iteration during the training phase, we randomly select one subject for training. 
Here, let the selected subject be the actor for a simplified explanation. 
The face region image $I_{act}$ is fed to the shared encoder $E_{shared}$ to produce the latent expression code $\gamma_{act}$ in the shared latent space. 
Then, the latent code $\gamma_{act}$ is fed into the decoder $D_{act}$ for the generation of the face image $I_{act}^{'}$, which will be optimized by the self-reconstruction loss function:
\begin{equation}
\mathcal{L}_{rec} = ||I_{act}^{'} - I_{act}||_{1} + \lambda_{ssim} SSIM(I_{act}^{'}, I_{act}),
\end{equation}
where $SSIM(\cdot)$ denotes Structure Similarity Index and $\lambda_{ssim}$ denotes the weight.

In the original Y-shaped network~\cite{yan2018video}, there is no restriction imposed on the learning of the shared latent space, which may lead to a separate distribution of the latent codes of input subjects.
However, to ensure the expression consistency of different subjects, the distributions of the shared latent code should overlap with each other as much as possible for the two input subjects.
Inspired by CycleGAN~\cite{zhu2017unpaired}, we leverage a similar cycle consistency loss to guarantee this. 
Specifically, the actor latent code $\gamma_{act}$ is fed into the avatar decoder $D_{ava}$ to generate the avatar image $I_{ava}^{*}$, which will be further fed into the shared encoder $E_{shared}$ to obtain a new avatar latent code $\gamma_{ava}^{*}$.
By taking $\gamma_{ava}^{*}$ as input, the actor decoder can generate another actor image $I_{act}^{*}$.
The cycle consistency loss is imposed on the newly generated latent code and images, which can be formulated as:
\begin{equation}
\begin{split}
\mathcal{L}_{cycle} = ||I_{act}^{*} - I_{act}||_{1} + \lambda_{ssim} SSIM(I_{act}^{*}, I_{act}) \\
+ \lambda_{code} ||\gamma_{ava}^{*} - \gamma_{act}||_{2}.
\end{split}
\end{equation}
Overall, the total loss function can be formulated as:
\begin{equation}
\mathcal{L} = \mathcal{L}_{rec} + \mathcal{L}_{cycle}.
\end{equation}

\subsection{Mapping MLP}

To enable the cross-identity reenactment of LatentAvatar, we additionally train a small mapping MLP to bridge the shared and person-specific latent codes $\theta$ and $\gamma$. Specifically, this mapping MLP maps the latent expression code in the shared latent space to the one in the person-specific latent space. 
In this way, the input data just needs to go through only one encoder $E_{shared}$ and a small mapping MLP to obtain the drive signal (i.e., the latent expression code $\theta$) for cross-identity reenactment, as shown in Fig.~\ref{fig:reenactment}. 

To train the mapping MLP, we only need the training data of the avatar. During training, we freeze all the parameters in the previously trained autoencoder module and the Y-shaped network and only optimize the parameters of the mapping MLP. Given a face image of the avatar $I_{ava}$, we encode it to two latent codes: the shared latent expression code $\gamma$ and the person-specific latent expression code $\theta$ through the corresponding encoders $E_{shared}$ and $E$, respectively. Then, the mapping MLP is used to map the shared latent code to the person-specific one. The loss function can be formulated as:
\begin{equation}
\mathcal{L} = ||E(I_{ava}) - Mapping(E_{shared}(I_{ava}))||_{2},
\end{equation}
where $Mapping(\cdot)$ denotes the mapping MLP. 
\section{Experiments}

\subsection{Implementation Details}

The training of our LatentAvatar requires the video data of different subjects.
Specifically, we use the videos from a public dataset MEAD~\cite{kaisiyuan2020mead} and our portrait video dataset collected by a hand-held mobile phone. For MEAD, we selected 3 representative subjects. Each video contains about 60,000 frames. For our dataset, we capture videos of 4 subjects. Each video contains about 10,000 frames. We trim 10\% fragment from the begining or the end of each video for evaluation and the remaining 90\% for training.
During the data preprocessing phase, all videos are resized to $512\times512$ resolution.
We also follow previous work~\cite{shanchuan2021robust, zhanghan2022modnet} to remove the background from each video frame to obtain ground truth portrait images for training. 

In our experiments, all input images of the encoder and the shared encoder are resized to the resolution of $128\times128$. 
Moreover, following the pre-processing strategy in DeepFaceLab~\cite{ivan2020deepfacelab}, we localize the face region using the detected 68 face landmarks so that the input images of the encoder mainly contain the valid face region.
The dimensions of the person-specific and shared latent expression code are set as 128 and 256, respectively. The resolutions of both the tri-plane features and the low-resolution feature map are set as $128\times128$ while their channel number is set as 32. 

During optimization, we use an Adam~\cite{diederik2015adam} optimizer.
The learning rate is set as $1 \times 10^{-4}$ for all learnable parameters. For ray sampling in volumetric rendering, we sample 64 points along each ray in each iteration. The latent head NeRF model is trained for 200,000 iterations with a batch size of 2, while the Y-shaped architecture and the mapping MLP are trained for 30,000 iterations with a batch size of 16. 
The weights of different loss terms are set as follows: $\lambda_{vgg}=0.1$, $\lambda_{lr}=0.1$, $\lambda_{mask}=0.1$, $\lambda_{ssim}=0.1$ and $\lambda_{code}=1 \times 10^{-4}$.

\subsection{Results and Comparisons}

\begin{figure*}[ht]
  \centering
  \includegraphics[width=1.0\linewidth]{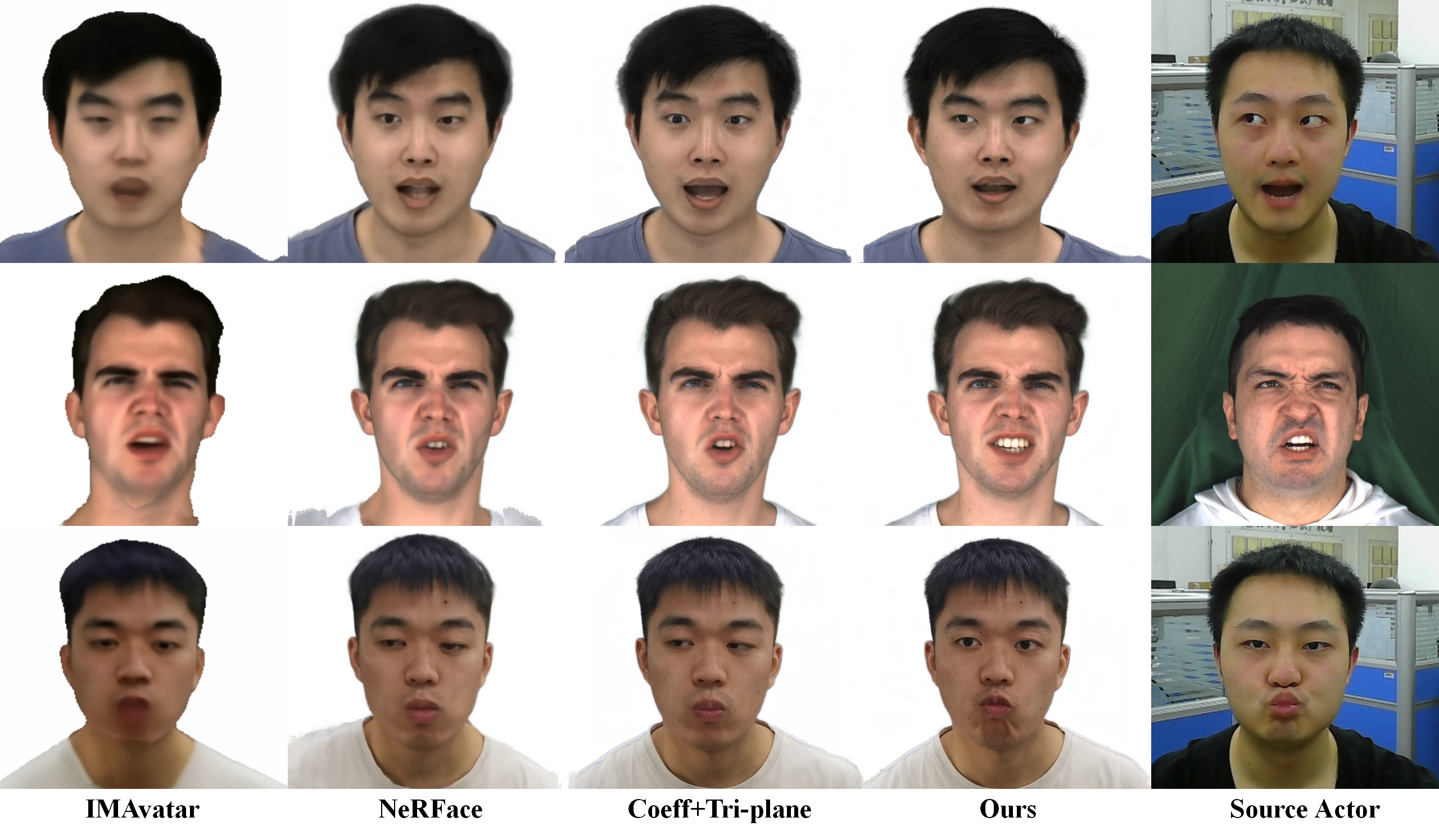}
  \vspace{-3mm}
  \caption{Qualitative comparisons of different methods on the cross-identity reenactment task. From left to right: IMavata, NeRFace, Coeff+Tri-plane baseline and Ours. Our method  can accurately transfer eye movement and tooth grinning and remain robust in some exaggerated expressions.}
  \vspace{-3mm}
  \label{fig:cross-identity reenactment}
\end{figure*}

The key idea of our LatentAvatar is to use the latent expression code as the drive signal to animate the head avatar.
To validate its efficacy, we build an ablation baseline method named Coeff+Triplane. 
In Coeff+Triplane, the architecture of the head NeRF module is exactly the same as our method, i.e., using the tri-plane representation and the super-resolution module to improve the quality of the generated images.
Just as NeRFace~\cite{gafni2021dynamic}, Coeff+Triplane directly uses the 3DMM expression coefficients as the drive signal, which is the only difference in comparison with our method.

\subsubsection{Self Reenactment}

We conduct qualitative and quantitative comparisons between our method, Coeff+Tri-plane, and other two state-of-the-art methods, i.e., IMavatar~\cite{zheng2022imavatar}, and NeRFace~\cite{gafni2021dynamic}, on our dataset. IMavatar reconstructs an implicit represented~\cite{yariv2020multiview} head avatar based on FLAME model~\cite{li2017learning}, while NeRFace reconstructs a dynamic NeRF represented head avatar with 3DMM expression coefficients as the drive signal.

The self reenactment results of different methods are shown in Fig.~\ref{fig:self reenactment}.
We can see that the accuracy of the template fitting is crucial for IMavatar due to the reliance on face templates in its reconstruction process.
The performance of IMavatar is inferior to our method, especially on our newly collected video data. 
Our video data contains a large number of challenging and dramatic expressions, resulting in inaccurate landmark detection results and the failure of template fitting. 
As a result, the texture of the reconstructed avatar of IMavatar tends to be much more blurred. 
Though NeRFace and Coeff+Tri-plane do not use the geometry or texture of the face template, they still rely on the 3DMM expression coefficients. 
As shown in Fig.~\ref{fig:self reenactment}, their reconstructed head avatar is not able to produce detailed wrinkles (see the first row) or eye movements (see the second row) since these subtle expressions cannot be explicitly represented by the face template. 
In contrast, the drive signal of our method is the latent expression code extracted from the input face image directly.
Such a learnable expression code is able to capture high-frequency and detailed expression information, which goes far beyond the expression ability of existing face templates.


For quantitative evaluation, all the methods are quantitatively evaluated by calculating Mean Square Error (MSE), Peak Signal-to-Noise Ratio (PSNR), Structure Similarity Index (SSIM) and Learned Perceptual Image Patch Similarity (LPIPS)~\cite{zhang2018the} between the ground truth image and the generated image on the evaluation data. 
During the evaluation, we select 6 avatars to perform self reenactment and calculate the mean values of the evaluation results.
The numerical results are reported in Tab.~\ref{tab:evaluation} for comparisons. 
We can see that our method convincingly outperforms all other state-of-the-art methods. 
\begin{table}[ht]
\centering
\caption{Quantitative evaluation results of our method, Coeff+Tri-plane (baseline), IMavatar~\cite{zheng2022imavatar}, and NeRFace~\cite{gafni2021dynamic}.}
\begin{tabular}{c|c|c|c|c}
\hline
Method             & MSE$\times10^{-3}$ $\downarrow$ & PSNR $\uparrow$    & SSIM $\uparrow$    & LPIPS $\downarrow$   \\
\hline
\hline
IMavatar           & 6.89                            & 21.79              & 0.871              & 0.209                \\
NeRFace            & 3.39                            & 25.64              & 0.903              & 0.135                \\
Coeff+Tri-plane    & 2.70                            & 27.00              & 0.917              & 0.049                \\
Ours               & \textbf{2.61}                   & \textbf{27.61}     & \textbf{0.919}     & \textbf{0.048}       \\
\hline
\end{tabular} 
\vspace{-1mm}
\label{tab:evaluation}
\end{table}

\subsubsection{Cross-identity Reenactment}
We show the cross-identity reenactment results of different methods in Fig.~\ref{fig:cross-identity reenactment}. 
For NeRFace and Coeff+Tri-plane, we can observe that these two methods may produce unreasonable results during the expression transfer between different subjects.
These can be explained by the fact that the identity coefficients and the expression coefficients of the face templates are easy to be coupled with each other.
In addition, these template-based methods cannot produce avatar images with challenging expressions such as pouting when the corresponding landmark shape is difficult to be detected.
As shown in Fig.~\ref{fig:cross-identity reenactment}, our method performs well under challenging cases and even captures the details with tooth baring (see the 2nd row), while the template-based methods suffer from serious artifacts when the expression coefficients are out of the distributions in the training dataset (see the 3rd row).
The success of our method can be attributed to the expression codes learned in the person-specific and shared latent space.
Both of these two latent codes and their mapping are learned in an end-to-end and self-supervised manner, enabling the capture of subtle expressions and their correspondences between different subjects.
Moreover, our method does not rely on template fitting and hence enjoys both of expressiveness and stability.

\begin{figure}[ht]
  \centering
  \includegraphics[width=\linewidth]{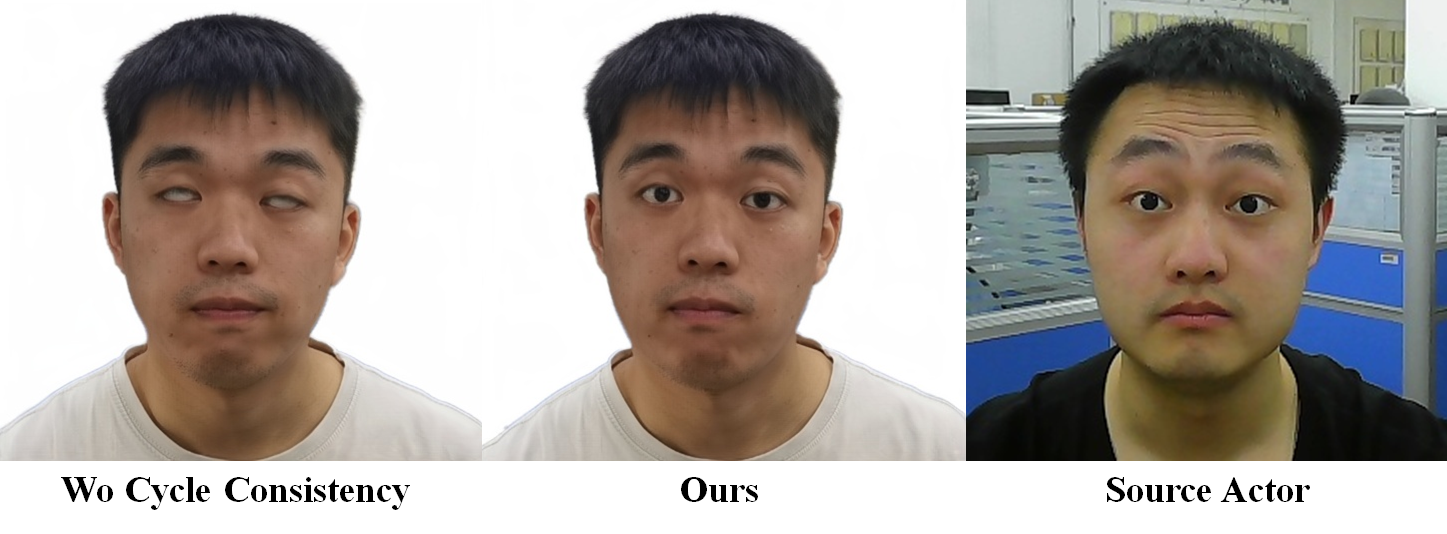}
  \caption{Ablation study on cycle consistency loss}
  \label{fig:ablation}
  \vspace{-2mm}
\end{figure}

\subsubsection{Ablation Study on Cycle Consistency Loss}

We compared our training pipeline with and without cycle consistency loss in cross-identity reenactment experiments. Qualitative results are shown in Fig.~\ref{fig:ablation}. Since there is no cycle consistency loss to constraint, the latent codes are more diffuse in the shared latent space, resulting in the expression consistency reducing significantly.

\section{Conclusion}

In this work, we have proposed LatentAvatar, an expressive NeRF-based head avatar driven by latent expression codes.
LatentAvatar leverages a latent head NeRF and a Y-shaped network to learn the latent expression code in the person-specific and shared space respectively.
These two types of latent expression code are further bridged by a mapping MLP to achieve cross-identity reenactment.
Experimental results have demonstrated the capability of our method to capture detailed expressions and subtle movements of the teeth and eyeballs, which shows significant improvement over previous solutions.
Moreover, LatentAvatar gets rid of the template tracking issues and hence is more robust and stable to challenging expressions.
We believe that the combination of NeRF representations and the learned latent expression code is a promising direction to achieve lightweight and expressive head avatar reconstruction and reenactment.

\begin{figure}[ht]
  \centering
  \vspace{-2mm}
  \includegraphics[width=\linewidth]{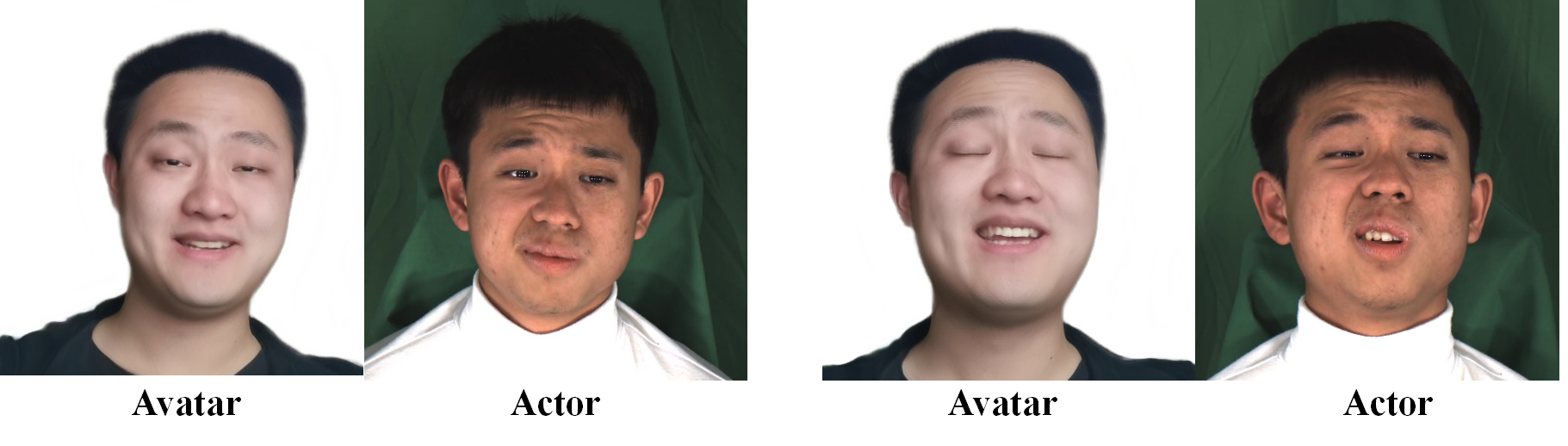}
  \caption{Failure cases when there are distinct differences between the appearance and expression distribution of the two identities.}
  \label{fig:failure}
  \vspace{-2mm}
\end{figure}

\section{Discussion}
\textbf{Ethical Considerations.} Our method can synthesize photo-realistic fake portrait videos, but also brings a serious problem that it can be used to spread false information, manipulate public opinion, and undermine trust in media, leading to significant societal harm. 
Therefore, an efficient and accurate method for discriminating forgery is the most worthy of consideration.

\textbf{Limitation.} Despite the expressive results, there are two main limitations of our method.
First, our method requires the learning of latent codes in person-specific and shared spaces, which implies that our method needs additional training on the video of the actor for the task of cross-identity reenactment.
Second, when there are distinct differences between the appearance or expression distribution of the two identities, the shared latent code is inclined to contain individual appearance information, which may lead to the erroneous mapping of expressions. Two failure cases are shown in Fig.~\ref{fig:failure} for illustration.

\textbf{Future Work.} In future work, to tackle the above two issues, we may adopt meta-learning techniques and leverage existing large-scale video datasets to enhance the generalization of our solution to diverse facial appearances and expressions.

\begin{acks}
This paper is supported by National Key R\&D Program of China (2022YFF0902200), the NSFC project No.62125107 and No.61827805.
\end{acks}
\bibliographystyle{ACM-Reference-Format}
\bibliography{base}


\begin{thebibliography}{92}


\ifx \showCODEN    \undefined \def \showCODEN     #1{\unskip}     \fi
\ifx \showDOI      \undefined \def \showDOI       #1{#1}\fi
\ifx \showISBNx    \undefined \def \showISBNx     #1{\unskip}     \fi
\ifx \showISBNxiii \undefined \def \showISBNxiii  #1{\unskip}     \fi
\ifx \showISSN     \undefined \def \showISSN      #1{\unskip}     \fi
\ifx \showLCCN     \undefined \def \showLCCN      #1{\unskip}     \fi
\ifx \shownote     \undefined \def \shownote      #1{#1}          \fi
\ifx \showarticletitle \undefined \def \showarticletitle #1{#1}   \fi
\ifx \showURL      \undefined \def \showURL       {\relax}        \fi
\providecommand\bibfield[2]{#2}
\providecommand\bibinfo[2]{#2}
\providecommand\natexlab[1]{#1}
\providecommand\showeprint[2][]{arXiv:#2}

\bibitem[Athar et~al\mbox{.}(2023)]%
        {athar2021flame}
\bibfield{author}{\bibinfo{person}{ShahRukh Athar}, \bibinfo{person}{Zhixin
  Shu}, {and} \bibinfo{person}{Dimitris Samaras}.}
  \bibinfo{year}{2023}\natexlab{}.
\newblock \showarticletitle{Flame-in-nerf: Neural control of radiance fields
  for free view face animation}. In \bibinfo{booktitle}{\emph{IEEE 17th
  International Conference on Automatic Face and Gesture Recognition (FG)}}.
  \bibinfo{pages}{1--8}.
\newblock


\bibitem[Athar et~al\mbox{.}(2022)]%
        {athar2022rignerf}
\bibfield{author}{\bibinfo{person}{ShahRukh Athar}, \bibinfo{person}{Zexiang
  Xu}, \bibinfo{person}{Kalyan Sunkavalli}, \bibinfo{person}{Eli Shechtman},
  {and} \bibinfo{person}{Zhixin Shu}.} \bibinfo{year}{2022}\natexlab{}.
\newblock \showarticletitle{RigNeRF: Fully Controllable Neural 3D Portraits}.
  In \bibinfo{booktitle}{\emph{Proceedings of the IEEE/CVF Conference on
  Computer Vision and Pattern Recognition (CVPR)}}.
\newblock


\bibitem[Averbuch-Elor et~al\mbox{.}(2017)]%
        {avebuch2017bringing}
\bibfield{author}{\bibinfo{person}{Hadar Averbuch-Elor},
  \bibinfo{person}{Daniel Cohen-Or}, \bibinfo{person}{Johannes Kopf}, {and}
  \bibinfo{person}{Michael~F. Cohen}.} \bibinfo{year}{2017}\natexlab{}.
\newblock \showarticletitle{Bringing Portraits to Life}.
\newblock \bibinfo{journal}{\emph{ACM Trans. Graph.}} \bibinfo{volume}{36},
  \bibinfo{number}{6}, Article \bibinfo{articleno}{196} (\bibinfo{date}{nov}
  \bibinfo{year}{2017}), \bibinfo{numpages}{13}~pages.
\newblock
\showISSN{0730-0301}


\bibitem[Bi et~al\mbox{.}(2021)]%
        {bi2021deep}
\bibfield{author}{\bibinfo{person}{Sai Bi}, \bibinfo{person}{Stephen Lombardi},
  \bibinfo{person}{Shunsuke Saito}, \bibinfo{person}{Tomas Simon},
  \bibinfo{person}{Shih-En Wei}, \bibinfo{person}{Kevyn Mcphail},
  \bibinfo{person}{Ravi Ramamoorthi}, \bibinfo{person}{Yaser Sheikh}, {and}
  \bibinfo{person}{Jason Saragih}.} \bibinfo{year}{2021}\natexlab{}.
\newblock \showarticletitle{Deep Relightable Appearance Models for Animatable
  Faces}.
\newblock \bibinfo{journal}{\emph{ACM Trans. Graph.}} \bibinfo{volume}{40},
  \bibinfo{number}{4}, Article \bibinfo{articleno}{89} (\bibinfo{date}{jul}
  \bibinfo{year}{2021}), \bibinfo{numpages}{15}~pages.
\newblock


\bibitem[Cao et~al\mbox{.}(2021)]%
        {cao2021real}
\bibfield{author}{\bibinfo{person}{Chen Cao}, \bibinfo{person}{Vasu Agrawal},
  \bibinfo{person}{Fernando De~La~Torre}, \bibinfo{person}{Lele Chen},
  \bibinfo{person}{Jason Saragih}, \bibinfo{person}{Tomas Simon}, {and}
  \bibinfo{person}{Yaser Sheikh}.} \bibinfo{year}{2021}\natexlab{}.
\newblock \showarticletitle{Real-Time 3D Neural Facial Animation from Binocular
  Video}.
\newblock \bibinfo{journal}{\emph{ACM Trans. Graph.}} \bibinfo{volume}{40},
  \bibinfo{number}{4}, Article \bibinfo{articleno}{87} (\bibinfo{date}{jul}
  \bibinfo{year}{2021}), \bibinfo{numpages}{17}~pages.
\newblock


\bibitem[Cao et~al\mbox{.}(2015)]%
        {cao2015real}
\bibfield{author}{\bibinfo{person}{Chen Cao}, \bibinfo{person}{Derek Bradley},
  \bibinfo{person}{Kun Zhou}, {and} \bibinfo{person}{Thabo Beeler}.}
  \bibinfo{year}{2015}\natexlab{}.
\newblock \showarticletitle{Real-Time High-Fidelity Facial Performance
  Capture}.
\newblock \bibinfo{journal}{\emph{ACM Trans. Graph.}} \bibinfo{volume}{34},
  \bibinfo{number}{4}, Article \bibinfo{articleno}{46} (\bibinfo{date}{jul}
  \bibinfo{year}{2015}), \bibinfo{numpages}{9}~pages.
\newblock
\showISSN{0730-0301}


\bibitem[Cao et~al\mbox{.}(2014a)]%
        {cao2014displaced}
\bibfield{author}{\bibinfo{person}{Chen Cao}, \bibinfo{person}{Qiming Hou},
  {and} \bibinfo{person}{Kun Zhou}.} \bibinfo{year}{2014}\natexlab{a}.
\newblock \showarticletitle{Displaced Dynamic Expression Regression for
  Real-Time Facial Tracking and Animation}.
\newblock \bibinfo{journal}{\emph{ACM Trans. Graph.}} \bibinfo{volume}{33},
  \bibinfo{number}{4}, Article \bibinfo{articleno}{43} (\bibinfo{date}{jul}
  \bibinfo{year}{2014}), \bibinfo{numpages}{10}~pages.
\newblock
\showISSN{0730-0301}


\bibitem[Cao et~al\mbox{.}(2022)]%
        {cao2022authentic}
\bibfield{author}{\bibinfo{person}{Chen Cao}, \bibinfo{person}{Tomas Simon},
  \bibinfo{person}{Jin~Kyu Kim}, \bibinfo{person}{Gabe Schwartz},
  \bibinfo{person}{Michael Zollhoefer}, \bibinfo{person}{Shun-Suke Saito},
  \bibinfo{person}{Stephen Lombardi}, \bibinfo{person}{Shih-En Wei},
  \bibinfo{person}{Danielle Belko}, \bibinfo{person}{Shoou-I Yu},
  \bibinfo{person}{Yaser Sheikh}, {and} \bibinfo{person}{Jason Saragih}.}
  \bibinfo{year}{2022}\natexlab{}.
\newblock \showarticletitle{Authentic Volumetric Avatars from a Phone Scan}.
\newblock \bibinfo{journal}{\emph{ACM Trans. Graph.}} \bibinfo{volume}{41},
  \bibinfo{number}{4}, Article \bibinfo{articleno}{163} (\bibinfo{date}{jul}
  \bibinfo{year}{2022}), \bibinfo{numpages}{19}~pages.
\newblock
\showISSN{0730-0301}


\bibitem[Cao et~al\mbox{.}(2013)]%
        {cao20133d}
\bibfield{author}{\bibinfo{person}{Chen Cao}, \bibinfo{person}{Yanlin Weng},
  \bibinfo{person}{Stephen Lin}, {and} \bibinfo{person}{Kun Zhou}.}
  \bibinfo{year}{2013}\natexlab{}.
\newblock \showarticletitle{3D Shape Regression for Real-Time Facial
  Animation}.
\newblock \bibinfo{journal}{\emph{ACM Trans. Graph.}} \bibinfo{volume}{32},
  \bibinfo{number}{4}, Article \bibinfo{articleno}{41} (\bibinfo{date}{jul}
  \bibinfo{year}{2013}), \bibinfo{numpages}{10}~pages.
\newblock
\showISSN{0730-0301}


\bibitem[Cao et~al\mbox{.}(2014b)]%
        {cao2014facewarehouse}
\bibfield{author}{\bibinfo{person}{Chen Cao}, \bibinfo{person}{Yanlin Weng},
  \bibinfo{person}{Shun Zhou}, \bibinfo{person}{Y. Tong}, {and}
  \bibinfo{person}{Kun Zhou}.} \bibinfo{year}{2014}\natexlab{b}.
\newblock \showarticletitle{FaceWarehouse: A 3D Facial Expression Database for
  Visual Computing}. In \bibinfo{booktitle}{\emph{IEEE Transactions on
  Visualization and Computer Graphics}}, Vol.~\bibinfo{volume}{20}.
  \bibinfo{pages}{413--425}.
\newblock


\bibitem[Cao et~al\mbox{.}(2016)]%
        {cao2016real}
\bibfield{author}{\bibinfo{person}{Chen Cao}, \bibinfo{person}{Hongzhi Wu},
  \bibinfo{person}{Yanlin Weng}, \bibinfo{person}{Tianjia Shao}, {and}
  \bibinfo{person}{Kun Zhou}.} \bibinfo{year}{2016}\natexlab{}.
\newblock \showarticletitle{Real-Time Facial Animation with Image-Based Dynamic
  Avatars}.
\newblock \bibinfo{journal}{\emph{ACM Trans. Graph.}} \bibinfo{volume}{35},
  \bibinfo{number}{4}, Article \bibinfo{articleno}{126} (\bibinfo{date}{jul}
  \bibinfo{year}{2016}), \bibinfo{numpages}{12}~pages.
\newblock
\showISSN{0730-0301}


\bibitem[Chan et~al\mbox{.}(2022)]%
        {chan2022efficient}
\bibfield{author}{\bibinfo{person}{Eric~R. Chan}, \bibinfo{person}{Connor~Z.
  Lin}, \bibinfo{person}{Matthew~A. Chan}, \bibinfo{person}{Koki Nagano},
  \bibinfo{person}{Boxiao Pan}, \bibinfo{person}{Shalini~De Mello},
  \bibinfo{person}{Orazio Gallo}, \bibinfo{person}{Leonidas Guibas},
  \bibinfo{person}{Jonathan Tremblay}, \bibinfo{person}{Sameh Khamis},
  \bibinfo{person}{Tero Karras}, {and} \bibinfo{person}{Gordon Wetzstein}.}
  \bibinfo{year}{2022}\natexlab{}.
\newblock \showarticletitle{Efficient Geometry-aware {3D} Generative
  Adversarial Networks}. In \bibinfo{booktitle}{\emph{Proceedings of the
  IEEE/CVF Conference on Computer Vision and Pattern Recognition (CVPR)}}.
  \bibinfo{pages}{16102--16112}.
\newblock


\bibitem[Chen et~al\mbox{.}(2021)]%
        {chen2021mvsnerf}
\bibfield{author}{\bibinfo{person}{Anpei Chen}, \bibinfo{person}{Zexiang Xu},
  \bibinfo{person}{Fuqiang Zhao}, \bibinfo{person}{Xiaoshuai Zhang},
  \bibinfo{person}{Fanbo Xiang}, \bibinfo{person}{Jingyi Yu}, {and}
  \bibinfo{person}{Hao Su}.} \bibinfo{year}{2021}\natexlab{}.
\newblock \showarticletitle{Mvsnerf: Fast generalizable radiance field
  reconstruction from multi-view stereo}. In
  \bibinfo{booktitle}{\emph{Proceedings of the IEEE/CVF International
  Conference on Computer Vision}}. \bibinfo{pages}{14124--14133}.
\newblock


\bibitem[Chen et~al\mbox{.}(2020)]%
        {chen2020puppeteergan}
\bibfield{author}{\bibinfo{person}{Zhuo Chen}, \bibinfo{person}{Chaoyue Wang},
  \bibinfo{person}{Bo Yuan}, {and} \bibinfo{person}{Dacheng Tao}.}
  \bibinfo{year}{2020}\natexlab{}.
\newblock \showarticletitle{PuppeteerGAN: Arbitrary Portrait Animation With
  Semantic-Aware Appearance Transformation}. In
  \bibinfo{booktitle}{\emph{Proceedings of the IEEE/CVF Conference on Computer
  Vision and Pattern Recognition (CVPR)}}. \bibinfo{pages}{13515--13524}.
\newblock


\bibitem[Chu et~al\mbox{.}(2020)]%
        {chu2020expressive}
\bibfield{author}{\bibinfo{person}{Hang Chu}, \bibinfo{person}{Shugao Ma},
  \bibinfo{person}{Fernando Torre}, \bibinfo{person}{Sanja Fidler}, {and}
  \bibinfo{person}{Yaser Sheikh}.} \bibinfo{year}{2020}\natexlab{}.
\newblock \showarticletitle{Expressive Telepresence via Modular Codec Avatars}.
  In \bibinfo{booktitle}{\emph{Proceedings of the Proceedings of the European
  Conference on Computer Vision (ECCV)}}. \bibinfo{pages}{330--345}.
\newblock


\bibitem[Dale et~al\mbox{.}(2011)]%
        {dale2011video}
\bibfield{author}{\bibinfo{person}{Kevin Dale}, \bibinfo{person}{Kalyan
  Sunkavalli}, \bibinfo{person}{Micah~K. Johnson}, \bibinfo{person}{Daniel
  Vlasic}, \bibinfo{person}{Wojciech Matusik}, {and} \bibinfo{person}{Hanspeter
  Pfister}.} \bibinfo{year}{2011}\natexlab{}.
\newblock \showarticletitle{Video Face Replacement}.
\newblock \bibinfo{journal}{\emph{ACM Trans. Graph.}} \bibinfo{volume}{30},
  \bibinfo{number}{6} (\bibinfo{date}{dec} \bibinfo{year}{2011}),
  \bibinfo{pages}{1–10}.
\newblock
\showISSN{0730-0301}


\bibitem[Deng et~al\mbox{.}(2019)]%
        {deng2019accurate}
\bibfield{author}{\bibinfo{person}{Yu Deng}, \bibinfo{person}{Jiaolong Yang},
  \bibinfo{person}{Sicheng Xu}, \bibinfo{person}{Dong Chen},
  \bibinfo{person}{Yunde Jia}, {and} \bibinfo{person}{Xin Tong}.}
  \bibinfo{year}{2019}\natexlab{}.
\newblock \showarticletitle{Accurate 3D Face Reconstruction With
  Weakly-Supervised Learning: From Single Image to Image Set}. In
  \bibinfo{booktitle}{\emph{Proceedings of the IEEE/CVF Conference on Computer
  Vision and Pattern Recognition (CVPR) Workshops}}.
\newblock


\bibitem[Doukas et~al\mbox{.}(2020)]%
        {doukas2021head2head++}
\bibfield{author}{\bibinfo{person}{Michail~Christos Doukas},
  \bibinfo{person}{Mohammad~Rami Koujan}, \bibinfo{person}{Viktoriia
  Sharmanska}, \bibinfo{person}{Anastasios Roussos}, {and}
  \bibinfo{person}{Stefanos Zafeiriou}.} \bibinfo{year}{2020}\natexlab{}.
\newblock \showarticletitle{Head2Head++: Deep Facial Attributes Re-Targeting}.
\newblock \bibinfo{journal}{\emph{IEEE Transactions on Biometrics, Behavior,
  and Identity Science}}  \bibinfo{volume}{3} (\bibinfo{year}{2020}),
  \bibinfo{pages}{31--43}.
\newblock


\bibitem[Doukas et~al\mbox{.}(2021)]%
        {doukas2021headgan}
\bibfield{author}{\bibinfo{person}{Michail~Christos Doukas},
  \bibinfo{person}{Stefanos Zafeiriou}, {and} \bibinfo{person}{Viktoriia
  Sharmanska}.} \bibinfo{year}{2021}\natexlab{}.
\newblock \showarticletitle{HeadGAN: One-shot Neural Head Synthesis and
  Editing}. In \bibinfo{booktitle}{\emph{Proceedings of the IEEE/CVF
  International Conference on Computer Vision (ICCV)}}.
\newblock


\bibitem[Drobyshev et~al\mbox{.}(2022)]%
        {drobyshev2022megaportraits}
\bibfield{author}{\bibinfo{person}{Nikita Drobyshev}, \bibinfo{person}{Jenya
  Chelishev}, \bibinfo{person}{Taras Khakhulin}, \bibinfo{person}{Aleksei
  Ivakhnenko}, \bibinfo{person}{Victor Lempitsky}, {and} \bibinfo{person}{Egor
  Zakharov}.} \bibinfo{year}{2022}\natexlab{}.
\newblock \showarticletitle{MegaPortraits: One-shot Megapixel Neural Head
  Avatars}.
\newblock \bibinfo{journal}{\emph{In Proceedings of the 30th ACM International
  Conference on Multimedia}}.
\newblock


\bibitem[Fang et~al\mbox{.}(2022)]%
        {fang2022fast}
\bibfield{author}{\bibinfo{person}{Jiemin Fang}, \bibinfo{person}{Taoran Yi},
  \bibinfo{person}{Xinggang Wang}, \bibinfo{person}{Lingxi Xie},
  \bibinfo{person}{Xiaopeng Zhang}, \bibinfo{person}{Wenyu Liu},
  \bibinfo{person}{Matthias Nie\ss{}ner}, {and} \bibinfo{person}{Qi Tian}.}
  \bibinfo{year}{2022}\natexlab{}.
\newblock \showarticletitle{Fast Dynamic Radiance Fields with Time-Aware Neural
  Voxels}. In \bibinfo{booktitle}{\emph{SIGGRAPH Asia 2022 Conference Papers}}.
\newblock


\bibitem[Gafni et~al\mbox{.}(2021)]%
        {gafni2021dynamic}
\bibfield{author}{\bibinfo{person}{Guy Gafni}, \bibinfo{person}{Justus Thies},
  \bibinfo{person}{Michael Zollhofer}, {and} \bibinfo{person}{Matthias
  Niessner}.} \bibinfo{year}{2021}\natexlab{}.
\newblock \showarticletitle{Dynamic Neural Radiance Fields for Monocular 4D
  Facial Avatar Reconstruction}. In \bibinfo{booktitle}{\emph{Proceedings of
  the IEEE/CVF Conference on Computer Vision and Pattern Recognition (CVPR)}}.
  \bibinfo{pages}{8645--8654}.
\newblock


\bibitem[Gao et~al\mbox{.}(2022)]%
        {gao2022reconstructing}
\bibfield{author}{\bibinfo{person}{Xuan Gao}, \bibinfo{person}{Chenglai Zhong},
  \bibinfo{person}{Jun Xiang}, \bibinfo{person}{Yang Hong},
  \bibinfo{person}{Yudong Guo}, {and} \bibinfo{person}{Juyong Zhang}.}
  \bibinfo{year}{2022}\natexlab{}.
\newblock \showarticletitle{Reconstructing Personalized Semantic Facial NeRF
  Models From Monocular Video}.
\newblock \bibinfo{journal}{\emph{ACM Transactions on Graphics (Proceedings of
  SIGGRAPH Asia)}} \bibinfo{volume}{41}, \bibinfo{number}{6}
  (\bibinfo{year}{2022}).
\newblock


\bibitem[Geng et~al\mbox{.}(2018)]%
        {geng2018warp}
\bibfield{author}{\bibinfo{person}{Jiahao Geng}, \bibinfo{person}{Tianjia
  Shao}, \bibinfo{person}{Youyi Zheng}, \bibinfo{person}{Yanlin Weng}, {and}
  \bibinfo{person}{Kun Zhou}.} \bibinfo{year}{2018}\natexlab{}.
\newblock \showarticletitle{Warp-Guided GANs for Single-Photo Facial
  Animation}.
\newblock \bibinfo{journal}{\emph{ACM Trans. Graph.}} \bibinfo{volume}{37},
  \bibinfo{number}{6}, Article \bibinfo{articleno}{231} (\bibinfo{date}{dec}
  \bibinfo{year}{2018}), \bibinfo{numpages}{12}~pages.
\newblock
\showISSN{0730-0301}


\bibitem[Gerig et~al\mbox{.}(2017)]%
        {gerig2018morphable}
\bibfield{author}{\bibinfo{person}{Thomas Gerig}, \bibinfo{person}{Andreas
  Forster}, \bibinfo{person}{Clemens Blumer}, \bibinfo{person}{Bernhard Egger},
  \bibinfo{person}{Marcel L{\"u}thi}, \bibinfo{person}{Sandro Sch{\"o}nborn},
  {and} \bibinfo{person}{Thomas Vetter}.} \bibinfo{year}{2017}\natexlab{}.
\newblock \showarticletitle{Morphable Face Models - An Open Framework}.
\newblock \bibinfo{journal}{\emph{2018 13th IEEE International Conference on
  Automatic Face \& Gesture Recognition (FG 2018)}}, \bibinfo{pages}{75--82}.
\newblock


\bibitem[Goodfellow et~al\mbox{.}(2014)]%
        {goodfellow2014generative}
\bibfield{author}{\bibinfo{person}{Ian Goodfellow}, \bibinfo{person}{Jean
  Pouget-Abadie}, \bibinfo{person}{Mehdi Mirza}, \bibinfo{person}{Bing Xu},
  \bibinfo{person}{David Warde-Farley}, \bibinfo{person}{Sherjil Ozair},
  \bibinfo{person}{Aaron Courville}, {and} \bibinfo{person}{Yoshua Bengio}.}
  \bibinfo{year}{2014}\natexlab{}.
\newblock \showarticletitle{Generative Adversarial Nets}. In
  \bibinfo{booktitle}{\emph{Conference on Neural Information Processing Systems
  (NeurIPS)}}, \bibfield{editor}{\bibinfo{person}{Z.~Ghahramani},
  \bibinfo{person}{M.~Welling}, \bibinfo{person}{C.~Cortes},
  \bibinfo{person}{N.~Lawrence}, {and} \bibinfo{person}{K.Q. Weinberger}}
  (Eds.), Vol.~\bibinfo{volume}{27}. \bibinfo{publisher}{Curran Associates,
  Inc.}
\newblock


\bibitem[Grassal et~al\mbox{.}(2022)]%
        {grassal2022neural}
\bibfield{author}{\bibinfo{person}{Philip-William Grassal},
  \bibinfo{person}{Malte Prinzler}, \bibinfo{person}{Titus Leistner},
  \bibinfo{person}{Carsten Rother}, \bibinfo{person}{Matthias Nie{\ss}ner},
  {and} \bibinfo{person}{Justus Thies}.} \bibinfo{year}{2022}\natexlab{}.
\newblock \showarticletitle{Neural Head Avatars From Monocular RGB Videos}. In
  \bibinfo{booktitle}{\emph{Proceedings of the IEEE/CVF Conference on Computer
  Vision and Pattern Recognition (CVPR)}}. \bibinfo{pages}{18632--18643}.
\newblock


\bibitem[Guo et~al\mbox{.}(2021)]%
        {guo2021ad}
\bibfield{author}{\bibinfo{person}{Yudong Guo}, \bibinfo{person}{Keyu Chen},
  \bibinfo{person}{Sen Liang}, \bibinfo{person}{Yong-Jin Liu},
  \bibinfo{person}{Hujun Bao}, {and} \bibinfo{person}{Juyong Zhang}.}
  \bibinfo{year}{2021}\natexlab{}.
\newblock \showarticletitle{AD-NeRF: Audio Driven Neural Radiance Fields for
  Talking Head Synthesis}. In \bibinfo{booktitle}{\emph{Proceedings of the
  IEEE/CVF International Conference on Computer Vision (ICCV)}}.
  \bibinfo{pages}{5764--5774}.
\newblock


\bibitem[Hong et~al\mbox{.}(2022)]%
        {hong2022headnerf}
\bibfield{author}{\bibinfo{person}{Yang Hong}, \bibinfo{person}{Bo Peng},
  \bibinfo{person}{Haiyao Xiao}, \bibinfo{person}{Ligang Liu}, {and}
  \bibinfo{person}{Juyong Zhang}.} \bibinfo{year}{2022}\natexlab{}.
\newblock \showarticletitle{HeadNeRF: A Real-Time NeRF-Based Parametric Head
  Model}. In \bibinfo{booktitle}{\emph{Proceedings of the IEEE/CVF Conference
  on Computer Vision and Pattern Recognition (CVPR)}}.
  \bibinfo{pages}{20374--20384}.
\newblock


\bibitem[Hu et~al\mbox{.}(2017)]%
        {hu2017avatar}
\bibfield{author}{\bibinfo{person}{Liwen Hu}, \bibinfo{person}{Shunsuke Saito},
  \bibinfo{person}{Lingyu Wei}, \bibinfo{person}{Koki Nagano},
  \bibinfo{person}{Jaewoo Seo}, \bibinfo{person}{Jens Fursund},
  \bibinfo{person}{Iman Sadeghi}, \bibinfo{person}{Carrie Sun},
  \bibinfo{person}{Yen-Chun Chen}, {and} \bibinfo{person}{Hao Li}.}
  \bibinfo{year}{2017}\natexlab{}.
\newblock \showarticletitle{Avatar Digitization from a Single Image for
  Real-Time Rendering}.
\newblock \bibinfo{journal}{\emph{ACM Trans. Graph.}} \bibinfo{volume}{36},
  \bibinfo{number}{6}, Article \bibinfo{articleno}{195} (\bibinfo{date}{nov}
  \bibinfo{year}{2017}), \bibinfo{numpages}{14}~pages.
\newblock
\showISSN{0730-0301}


\bibitem[Ichim et~al\mbox{.}(2015)]%
        {ichim2015dynamic}
\bibfield{author}{\bibinfo{person}{Alexandru~Eugen Ichim},
  \bibinfo{person}{Sofien Bouaziz}, {and} \bibinfo{person}{Mark Pauly}.}
  \bibinfo{year}{2015}\natexlab{}.
\newblock \showarticletitle{Dynamic 3D Avatar Creation from Hand-Held Video
  Input}.
\newblock \bibinfo{journal}{\emph{ACM Trans. Graph.}} \bibinfo{volume}{34},
  \bibinfo{number}{4}, Article \bibinfo{articleno}{45} (\bibinfo{date}{jul}
  \bibinfo{year}{2015}), \bibinfo{numpages}{14}~pages.
\newblock
\showISSN{0730-0301}


\bibitem[Karras et~al\mbox{.}(2021)]%
        {karras2021a}
\bibfield{author}{\bibinfo{person}{Tero Karras}, \bibinfo{person}{Samuli
  Laine}, {and} \bibinfo{person}{Timo Aila}.} \bibinfo{year}{2021}\natexlab{}.
\newblock \showarticletitle{A Style-Based Generator Architecture for Generative
  Adversarial Networks}.
\newblock \bibinfo{journal}{\emph{IEEE Transactions on Pattern Analysis and
  Machine Intelligence}} \bibinfo{volume}{43}, \bibinfo{number}{12}
  (\bibinfo{year}{2021}), \bibinfo{pages}{4217--4228}.
\newblock


\bibitem[Ke et~al\mbox{.}(2020)]%
        {zhanghan2022modnet}
\bibfield{author}{\bibinfo{person}{Zhanghan Ke}, \bibinfo{person}{Jiayu Sun},
  \bibinfo{person}{Kaican Li}, \bibinfo{person}{Qiong Yan}, {and}
  \bibinfo{person}{Rynson W.~H. Lau}.} \bibinfo{year}{2020}\natexlab{}.
\newblock \showarticletitle{MODNet: Real-Time Trimap-Free Portrait Matting via
  Objective Decomposition}. In \bibinfo{booktitle}{\emph{AAAI Conference on
  Artificial Intelligence}}.
\newblock


\bibitem[Khakhulin et~al\mbox{.}(2022)]%
        {Khakhulin2022realistic}
\bibfield{author}{\bibinfo{person}{Taras Khakhulin}, \bibinfo{person}{Vanessa
  Sklyarova}, \bibinfo{person}{Victor Lempitsky}, {and} \bibinfo{person}{Egor
  Zakharov}.} \bibinfo{year}{2022}\natexlab{}.
\newblock \showarticletitle{Realistic One-shot Mesh-based Head Avatars}. In
  \bibinfo{booktitle}{\emph{Proceedings of the European Conference on Computer
  Vision (ECCV)}}.
\newblock


\bibitem[Kim et~al\mbox{.}(2018)]%
        {kim2018deep}
\bibfield{author}{\bibinfo{person}{Hyeongwoo Kim}, \bibinfo{person}{Pablo
  Garrido}, \bibinfo{person}{Ayush Tewari}, \bibinfo{person}{Weipeng Xu},
  \bibinfo{person}{Justus Thies}, \bibinfo{person}{Matthias Niessner},
  \bibinfo{person}{Patrick P\'{e}rez}, \bibinfo{person}{Christian Richardt},
  \bibinfo{person}{Michael Zollh\"{o}fer}, {and} \bibinfo{person}{Christian
  Theobalt}.} \bibinfo{year}{2018}\natexlab{}.
\newblock \showarticletitle{Deep Video Portraits}.
\newblock \bibinfo{journal}{\emph{ACM Trans. Graph.}} \bibinfo{volume}{37},
  \bibinfo{number}{4}, Article \bibinfo{articleno}{163} (\bibinfo{date}{jul}
  \bibinfo{year}{2018}), \bibinfo{numpages}{14}~pages.
\newblock
\showISSN{0730-0301}


\bibitem[Kingma and Ba(2017)]%
        {diederik2015adam}
\bibfield{author}{\bibinfo{person}{Diederik~P. Kingma} {and}
  \bibinfo{person}{Jimmy Ba}.} \bibinfo{year}{2017}\natexlab{}.
\newblock \bibinfo{title}{Adam: A Method for Stochastic Optimization}.
\newblock
\newblock
\showeprint[arxiv]{1412.6980}~[cs.LG]


\bibitem[Korshunova et~al\mbox{.}(2017)]%
        {korshunova2017fast}
\bibfield{author}{\bibinfo{person}{Iryna Korshunova}, \bibinfo{person}{Wenzhe
  Shi}, \bibinfo{person}{Joni Dambre}, {and} \bibinfo{person}{Lucas Theis}.}
  \bibinfo{year}{2017}\natexlab{}.
\newblock \showarticletitle{Fast Face-Swap Using Convolutional Neural
  Networks}. In \bibinfo{booktitle}{\emph{Proceedings of the IEEE/CVF
  International Conference on Computer Vision (ICCV)}}.
  \bibinfo{pages}{3697--3705}.
\newblock


\bibitem[Koujan et~al\mbox{.}(2020)]%
        {koujan2020head2head}
\bibfield{author}{\bibinfo{person}{Mohammad~Rami Koujan},
  \bibinfo{person}{Michail~Christos Doukas}, \bibinfo{person}{Anastasios
  Roussos}, {and} \bibinfo{person}{Stefanos Zafeiriou}.}
  \bibinfo{year}{2020}\natexlab{}.
\newblock \showarticletitle{Head2Head: Video-based Neural Head Synthesis}.
\newblock \bibinfo{journal}{\emph{2020 15th IEEE International Conference on
  Automatic Face and Gesture Recognition (FG 2020)}}, \bibinfo{pages}{16--23}.
\newblock


\bibitem[Li et~al\mbox{.}(2012)]%
        {li2012a}
\bibfield{author}{\bibinfo{person}{Kai Li}, \bibinfo{person}{Feng Xu},
  \bibinfo{person}{Jue Wang}, \bibinfo{person}{Qionghai Dai}, {and}
  \bibinfo{person}{Yebin Liu}.} \bibinfo{year}{2012}\natexlab{}.
\newblock \showarticletitle{A data-driven approach for facial expression
  synthesis in video}. In \bibinfo{booktitle}{\emph{Proceedings of the IEEE/CVF
  Conference on Computer Vision and Pattern Recognition (CVPR)}}.
  \bibinfo{pages}{299--310}.
\newblock


\bibitem[Li et~al\mbox{.}(2017)]%
        {li2017learning}
\bibfield{author}{\bibinfo{person}{Tianye Li}, \bibinfo{person}{Timo Bolkart},
  \bibinfo{person}{Michael~J. Black}, \bibinfo{person}{Hao Li}, {and}
  \bibinfo{person}{Javier Romero}.} \bibinfo{year}{2017}\natexlab{}.
\newblock \showarticletitle{Learning a Model of Facial Shape and Expression
  from 4D Scans}.
\newblock \bibinfo{journal}{\emph{ACM Trans. Graph.}} \bibinfo{volume}{36},
  \bibinfo{number}{6}, Article \bibinfo{articleno}{194} (\bibinfo{date}{nov}
  \bibinfo{year}{2017}), \bibinfo{numpages}{17}~pages.
\newblock
\showISSN{0730-0301}


\bibitem[Lin et~al\mbox{.}(2022)]%
        {shanchuan2021robust}
\bibfield{author}{\bibinfo{person}{Shanchuan Lin}, \bibinfo{person}{Linjie
  Yang}, \bibinfo{person}{Imran Saleemi}, {and} \bibinfo{person}{Soumyadip
  Sengupta}.} \bibinfo{year}{2022}\natexlab{}.
\newblock \showarticletitle{Robust High-Resolution Video Matting With Temporal
  Guidance}. In \bibinfo{booktitle}{\emph{In Proceedings of the IEEE/CVF Winter
  Conference on Applications of Computer Vision (WACV)}}.
  \bibinfo{pages}{3132--3141}.
\newblock


\bibitem[Liu et~al\mbox{.}(2022)]%
        {liu2022semantic}
\bibfield{author}{\bibinfo{person}{Xian Liu}, \bibinfo{person}{Yinghao Xu},
  \bibinfo{person}{Qianyi Wu}, \bibinfo{person}{Hang Zhou},
  \bibinfo{person}{Wayne Wu}, {and} \bibinfo{person}{Bolei Zhou}.}
  \bibinfo{year}{2022}\natexlab{}.
\newblock \showarticletitle{Semantic-Aware Implicit Neural Audio-Driven Video
  Portrait Generation}. In \bibinfo{booktitle}{\emph{Proceedings of the
  European Conference on Computer Vision (ECCV)}}.
\newblock


\bibitem[Lombardi et~al\mbox{.}(2018)]%
        {lombardi2018deep}
\bibfield{author}{\bibinfo{person}{Stephen Lombardi}, \bibinfo{person}{Jason
  Saragih}, \bibinfo{person}{Tomas Simon}, {and} \bibinfo{person}{Yaser
  Sheikh}.} \bibinfo{year}{2018}\natexlab{}.
\newblock \showarticletitle{Deep Appearance Models for Face Rendering}.
\newblock \bibinfo{journal}{\emph{ACM Trans. Graph.}} \bibinfo{volume}{37},
  \bibinfo{number}{4}, Article \bibinfo{articleno}{68} (\bibinfo{date}{July}
  \bibinfo{year}{2018}), \bibinfo{numpages}{13}~pages.
\newblock
\showISSN{0730-0301}


\bibitem[Lombardi et~al\mbox{.}(2019)]%
        {lombardi2019neural}
\bibfield{author}{\bibinfo{person}{Stephen Lombardi}, \bibinfo{person}{Tomas
  Simon}, \bibinfo{person}{Jason Saragih}, \bibinfo{person}{Gabriel Schwartz},
  \bibinfo{person}{Andreas Lehrmann}, {and} \bibinfo{person}{Yaser Sheikh}.}
  \bibinfo{year}{2019}\natexlab{}.
\newblock \showarticletitle{Neural Volumes: Learning Dynamic Renderable Volumes
  from Images}.
\newblock \bibinfo{journal}{\emph{ACM Trans. Graph.}} \bibinfo{volume}{38},
  \bibinfo{number}{4}, Article \bibinfo{articleno}{65} (\bibinfo{date}{July}
  \bibinfo{year}{2019}), \bibinfo{numpages}{14}~pages.
\newblock


\bibitem[Lombardi et~al\mbox{.}(2021)]%
        {lombardi2021mixture}
\bibfield{author}{\bibinfo{person}{Stephen Lombardi}, \bibinfo{person}{Tomas
  Simon}, \bibinfo{person}{Gabriel Schwartz}, \bibinfo{person}{Michael
  Zollhoefer}, \bibinfo{person}{Yaser Sheikh}, {and} \bibinfo{person}{Jason
  Saragih}.} \bibinfo{year}{2021}\natexlab{}.
\newblock \showarticletitle{Mixture of Volumetric Primitives for Efficient
  Neural Rendering}.
\newblock \bibinfo{journal}{\emph{ACM Trans. Graph.}} \bibinfo{volume}{40},
  \bibinfo{number}{4}, Article \bibinfo{articleno}{59} (\bibinfo{date}{jul}
  \bibinfo{year}{2021}), \bibinfo{numpages}{13}~pages.
\newblock


\bibitem[Ma et~al\mbox{.}(2021)]%
        {ma2021pixel}
\bibfield{author}{\bibinfo{person}{Shugao Ma}, \bibinfo{person}{Tomas Simon},
  \bibinfo{person}{Jason Saragih}, \bibinfo{person}{Dawei Wang},
  \bibinfo{person}{Yuecheng Li}, \bibinfo{person}{Fernando~De La~Torre}, {and}
  \bibinfo{person}{Yaser Sheikh}.} \bibinfo{year}{2021}\natexlab{}.
\newblock \showarticletitle{Pixel Codec Avatars}. In
  \bibinfo{booktitle}{\emph{2021 IEEE/CVF Conference on Computer Vision and
  Pattern Recognition (CVPR)}}. \bibinfo{pages}{64--73}.
\newblock


\bibitem[Mescheder et~al\mbox{.}(2019)]%
        {occupancy2019mescheder}
\bibfield{author}{\bibinfo{person}{Lars Mescheder}, \bibinfo{person}{Michael
  Oechsle}, \bibinfo{person}{Michael Niemeyer}, \bibinfo{person}{Sebastian
  Nowozin}, {and} \bibinfo{person}{Andreas Geiger}.}
  \bibinfo{year}{2019}\natexlab{}.
\newblock \showarticletitle{Occupancy Networks: Learning 3D Reconstruction in
  Function Space}. In \bibinfo{booktitle}{\emph{Proceedings of the IEEE/CVF
  Conference on Computer Vision and Pattern Recognition (CVPR)}}.
\newblock


\bibitem[Mildenhall et~al\mbox{.}(2020)]%
        {mildenhall2020nerf}
\bibfield{author}{\bibinfo{person}{Ben Mildenhall}, \bibinfo{person}{Pratul~P.
  Srinivasan}, \bibinfo{person}{Matthew Tancik}, \bibinfo{person}{Jonathan~T.
  Barron}, \bibinfo{person}{Ravi Ramamoorthi}, {and} \bibinfo{person}{Ren Ng}.}
  \bibinfo{year}{2020}\natexlab{}.
\newblock \showarticletitle{NeRF: Representing Scenes as Neural Radiance Fields
  for View Synthesis}. In \bibinfo{booktitle}{\emph{Proceedings of the European
  Conference on Computer Vision (ECCV)}}.
\newblock


\bibitem[Moser et~al\mbox{.}(2021)]%
        {moser2021semi}
\bibfield{author}{\bibinfo{person}{Lucio Moser}, \bibinfo{person}{Chinyu
  Chien}, \bibinfo{person}{Mark Williams}, \bibinfo{person}{Jose Serra},
  \bibinfo{person}{Darren Hendler}, {and} \bibinfo{person}{Doug Roble}.}
  \bibinfo{year}{2021}\natexlab{}.
\newblock \showarticletitle{Semi-Supervised Video-Driven Facial Animation
  Transfer for Production}.
\newblock \bibinfo{journal}{\emph{ACM Trans. Graph.}} \bibinfo{volume}{40},
  \bibinfo{number}{6}, Article \bibinfo{articleno}{222} (\bibinfo{date}{dec}
  \bibinfo{year}{2021}), \bibinfo{numpages}{18}~pages.
\newblock
\showISSN{0730-0301}


\bibitem[M\"uller et~al\mbox{.}(2022)]%
        {mueller2022instant}
\bibfield{author}{\bibinfo{person}{Thomas M\"uller}, \bibinfo{person}{Alex
  Evans}, \bibinfo{person}{Christoph Schied}, {and} \bibinfo{person}{Alexander
  Keller}.} \bibinfo{year}{2022}\natexlab{}.
\newblock \showarticletitle{Instant Neural Graphics Primitives with a
  Multiresolution Hash Encoding}.
\newblock \bibinfo{journal}{\emph{ACM Trans. Graph.}} \bibinfo{volume}{41},
  \bibinfo{number}{4}, Article \bibinfo{articleno}{102} (\bibinfo{date}{July}
  \bibinfo{year}{2022}), \bibinfo{numpages}{15}~pages.
\newblock


\bibitem[Nagano et~al\mbox{.}(2018)]%
        {nagano2018pagan}
\bibfield{author}{\bibinfo{person}{Koki Nagano}, \bibinfo{person}{Jaewoo Seo},
  \bibinfo{person}{Jun Xing}, \bibinfo{person}{Lingyu Wei},
  \bibinfo{person}{Zimo Li}, \bibinfo{person}{Shunsuke Saito},
  \bibinfo{person}{Aviral Agarwal}, \bibinfo{person}{Jens Fursund}, {and}
  \bibinfo{person}{Hao Li}.} \bibinfo{year}{2018}\natexlab{}.
\newblock \showarticletitle{PaGAN: Real-Time Avatars Using Dynamic Textures}.
\newblock \bibinfo{journal}{\emph{ACM Trans. Graph.}} \bibinfo{volume}{37},
  \bibinfo{number}{6}, Article \bibinfo{articleno}{258} (\bibinfo{date}{dec}
  \bibinfo{year}{2018}), \bibinfo{numpages}{12}~pages.
\newblock
\showISSN{0730-0301}


\bibitem[Naruniec et~al\mbox{.}(2020)]%
        {naruniec2020high}
\bibfield{author}{\bibinfo{person}{Jacek Naruniec}, \bibinfo{person}{Leonhard
  Helminger}, \bibinfo{person}{Christopher Schroers}, {and}
  \bibinfo{person}{Romann~M. Weber}.} \bibinfo{year}{2020}\natexlab{}.
\newblock \showarticletitle{High-Resolution Neural Face Swapping for Visual
  Effects}.
\newblock \bibinfo{journal}{\emph{Computer Graphics Forum}}
  \bibinfo{volume}{39}, \bibinfo{number}{4} (\bibinfo{year}{2020}),
  \bibinfo{pages}{173--184}.
\newblock


\bibitem[Natsume et~al\mbox{.}(2018)]%
        {natsume2019fsnet}
\bibfield{author}{\bibinfo{person}{Ryota Natsume}, \bibinfo{person}{Tatsuya
  Yatagawa}, {and} \bibinfo{person}{Shigeo Morishima}.}
  \bibinfo{year}{2018}\natexlab{}.
\newblock \showarticletitle{FSNet: An Identity-Aware Generative Model for
  Image-based Face Swapping}. In \bibinfo{booktitle}{\emph{Asian Conference on
  Computer Vision}}.
\newblock


\bibitem[Nirkin et~al\mbox{.}(2019a)]%
        {nirkin2019fsgan}
\bibfield{author}{\bibinfo{person}{Yuval Nirkin}, \bibinfo{person}{Yosi
  Keller}, {and} \bibinfo{person}{Tal Hassner}.}
  \bibinfo{year}{2019}\natexlab{a}.
\newblock \showarticletitle{FSGAN: Subject Agnostic Face Swapping and
  Reenactment}. In \bibinfo{booktitle}{\emph{Proceedings of the IEEE/CVF
  International Conference on Computer Vision (ICCV)}}.
  \bibinfo{pages}{7183--7192}.
\newblock


\bibitem[Nirkin et~al\mbox{.}(2019b)]%
        {natsume2018rsgan}
\bibfield{author}{\bibinfo{person}{Yuval Nirkin}, \bibinfo{person}{Yosi
  Keller}, {and} \bibinfo{person}{Tal Hassner}.}
  \bibinfo{year}{2019}\natexlab{b}.
\newblock \showarticletitle{FSGAN: Subject Agnostic Face Swapping and
  Reenactment}. In \bibinfo{booktitle}{\emph{Proceedings of the IEEE/CVF
  International Conference on Computer Vision (ICCV)}}.
  \bibinfo{pages}{7183--7192}.
\newblock


\bibitem[Nirkin et~al\mbox{.}(2018)]%
        {nirkin2018on}
\bibfield{author}{\bibinfo{person}{Yuval Nirkin}, \bibinfo{person}{Iacopo
  Masi}, \bibinfo{person}{Anh Tran~Tuan}, \bibinfo{person}{Tal Hassner}, {and}
  \bibinfo{person}{Gerard Medioni}.} \bibinfo{year}{2018}\natexlab{}.
\newblock \showarticletitle{On Face Segmentation, Face Swapping, and Face
  Perception}. In \bibinfo{booktitle}{\emph{2018 13th IEEE International
  Conference on Automatic Face \& Gesture Recognition (FG 2018)}}.
  \bibinfo{pages}{98--105}.
\newblock


\bibitem[Olszewski et~al\mbox{.}(2017)]%
        {olszewski2017realistic}
\bibfield{author}{\bibinfo{person}{Kyle Olszewski}, \bibinfo{person}{Zimo Li},
  \bibinfo{person}{Chao Yang}, \bibinfo{person}{Yi Zhou},
  \bibinfo{person}{Ronald Yu}, \bibinfo{person}{Zeng Huang},
  \bibinfo{person}{Sitao Xiang}, \bibinfo{person}{Shunsuke Saito},
  \bibinfo{person}{Pushmeet Kohli}, {and} \bibinfo{person}{Hao Li}.}
  \bibinfo{year}{2017}\natexlab{}.
\newblock \showarticletitle{Realistic Dynamic Facial Textures From a Single
  Image Using GANs}. In \bibinfo{booktitle}{\emph{Proceedings of the IEEE/CVF
  International Conference on Computer Vision (ICCV)}}.
  \bibinfo{pages}{5439--5448}.
\newblock


\bibitem[Park et~al\mbox{.}(2019)]%
        {park2019deepsdf}
\bibfield{author}{\bibinfo{person}{Jeong~Joon Park}, \bibinfo{person}{Peter
  Florence}, \bibinfo{person}{Julian Straub}, \bibinfo{person}{Richard
  Newcombe}, {and} \bibinfo{person}{Steven Lovegrove}.}
  \bibinfo{year}{2019}\natexlab{}.
\newblock \showarticletitle{DeepSDF: Learning Continuous Signed Distance
  Functions for Shape Representation}. In \bibinfo{booktitle}{\emph{Proceedings
  of the IEEE/CVF Conference on Computer Vision and Pattern Recognition
  (CVPR)}}. \bibinfo{pages}{165--174}.
\newblock


\bibitem[Park et~al\mbox{.}(2021a)]%
        {park2021nerfies}
\bibfield{author}{\bibinfo{person}{Keunhong Park}, \bibinfo{person}{Utkarsh
  Sinha}, \bibinfo{person}{Jonathan~T Barron}, \bibinfo{person}{Sofien
  Bouaziz}, \bibinfo{person}{Dan~B Goldman}, \bibinfo{person}{Steven~M Seitz},
  {and} \bibinfo{person}{Ricardo Martin-Brualla}.}
  \bibinfo{year}{2021}\natexlab{a}.
\newblock \showarticletitle{Nerfies: Deformable neural radiance fields}. In
  \bibinfo{booktitle}{\emph{Proceedings of the IEEE/CVF International
  Conference on Computer Vision (ICCV)}}. \bibinfo{pages}{5845--5854}.
\newblock


\bibitem[Park et~al\mbox{.}(2021b)]%
        {park2021hypernerf}
\bibfield{author}{\bibinfo{person}{Keunhong Park}, \bibinfo{person}{Utkarsh
  Sinha}, \bibinfo{person}{Peter Hedman}, \bibinfo{person}{Jonathan~T. Barron},
  \bibinfo{person}{Sofien Bouaziz}, \bibinfo{person}{Dan~B Goldman},
  \bibinfo{person}{Ricardo Martin-Brualla}, {and} \bibinfo{person}{Steven~M.
  Seitz}.} \bibinfo{year}{2021}\natexlab{b}.
\newblock \showarticletitle{HyperNeRF: A Higher-Dimensional Representation for
  Topologically Varying Neural Radiance Fields}.
\newblock \bibinfo{journal}{\emph{ACM Trans. Graph.}} \bibinfo{volume}{40},
  \bibinfo{number}{6}, Article \bibinfo{articleno}{238} (\bibinfo{date}{dec}
  \bibinfo{year}{2021}).
\newblock


\bibitem[Perov et~al\mbox{.}(2021)]%
        {ivan2020deepfacelab}
\bibfield{author}{\bibinfo{person}{Ivan Perov}, \bibinfo{person}{Daiheng Gao},
  \bibinfo{person}{Nikolay Chervoniy}, \bibinfo{person}{Kunlin Liu},
  \bibinfo{person}{Sugasa Marangonda}, \bibinfo{person}{Chris Umé},
  \bibinfo{person}{Mr. Dpfks}, \bibinfo{person}{Carl~Shift Facenheim},
  \bibinfo{person}{Luis RP}, \bibinfo{person}{Jian Jiang},
  \bibinfo{person}{Sheng Zhang}, \bibinfo{person}{Pingyu Wu},
  \bibinfo{person}{Bo Zhou}, {and} \bibinfo{person}{Weiming Zhang}.}
  \bibinfo{year}{2021}\natexlab{}.
\newblock \bibinfo{title}{DeepFaceLab: Integrated, flexible and extensible
  face-swapping framework}.
\newblock
\newblock
\showeprint[arxiv]{2005.05535}~[cs.CV]


\bibitem[Ren et~al\mbox{.}(2021)]%
        {ren2021pirenderer}
\bibfield{author}{\bibinfo{person}{Yurui Ren}, \bibinfo{person}{Ge Li},
  \bibinfo{person}{Yuanqi Chen}, \bibinfo{person}{Thomas~H. Li}, {and}
  \bibinfo{person}{Shan Liu}.} \bibinfo{year}{2021}\natexlab{}.
\newblock \showarticletitle{PIRenderer: Controllable Portrait Image Generation
  via Semantic Neural Rendering}. In \bibinfo{booktitle}{\emph{Proceedings of
  the IEEE/CVF International Conference on Computer Vision (ICCV)}}.
  \bibinfo{pages}{13759--13768}.
\newblock


\bibitem[Siarohin et~al\mbox{.}(2019)]%
        {siarohin2019first}
\bibfield{author}{\bibinfo{person}{Aliaksandr Siarohin},
  \bibinfo{person}{Stéphane Lathuilière}, \bibinfo{person}{Sergey Tulyakov},
  \bibinfo{person}{Elisa Ricci}, {and} \bibinfo{person}{Nicu Sebe}.}
  \bibinfo{year}{2019}\natexlab{}.
\newblock \showarticletitle{First Order Motion Model for Image Animation}. In
  \bibinfo{booktitle}{\emph{Conference on Neural Information Processing Systems
  (NeurIPS)}}.
\newblock


\bibitem[Sun et~al\mbox{.}(2022a)]%
        {sun2022ide}
\bibfield{author}{\bibinfo{person}{Jingxiang Sun}, \bibinfo{person}{Xuan Wang},
  \bibinfo{person}{Yichun Shi}, \bibinfo{person}{Lizhen Wang},
  \bibinfo{person}{Jue Wang}, {and} \bibinfo{person}{Yebin Liu}.}
  \bibinfo{year}{2022}\natexlab{a}.
\newblock \showarticletitle{IDE-3D: Interactive Disentangled Editing for
  High-Resolution 3D-aware Portrait Synthesis}.
\newblock \bibinfo{journal}{\emph{ACM Transactions on Graphics (TOG)}}
  \bibinfo{volume}{41}, \bibinfo{number}{6}, Article \bibinfo{articleno}{270}
  (\bibinfo{year}{2022}), \bibinfo{numpages}{10}~pages.
\newblock


\bibitem[Sun et~al\mbox{.}(2023)]%
        {sun2023next3d}
\bibfield{author}{\bibinfo{person}{Jingxiang Sun}, \bibinfo{person}{Xuan Wang},
  \bibinfo{person}{Lizhen Wang}, \bibinfo{person}{Xiaoyu Li},
  \bibinfo{person}{Yong Zhang}, \bibinfo{person}{Hongwen Zhang}, {and}
  \bibinfo{person}{Yebin Liu}.} \bibinfo{year}{2023}\natexlab{}.
\newblock \showarticletitle{Next3D: Generative Neural Texture Rasterization for
  3D-Aware Head Avatars}. In \bibinfo{booktitle}{\emph{Proceedings of the
  IEEE/CVF Conference on Computer Vision and Pattern Recognition (CVPR)}}.
\newblock


\bibitem[Sun et~al\mbox{.}(2022b)]%
        {sun2021fenerf}
\bibfield{author}{\bibinfo{person}{Jingxiang Sun}, \bibinfo{person}{Xuan Wang},
  \bibinfo{person}{Yong Zhang}, \bibinfo{person}{Xiaoyu Li},
  \bibinfo{person}{Qi Zhang}, \bibinfo{person}{Yebin Liu}, {and}
  \bibinfo{person}{Jue Wang}.} \bibinfo{year}{2022}\natexlab{b}.
\newblock \showarticletitle{FENeRF: Face Editing in Neural Radiance Fields}. In
  \bibinfo{booktitle}{\emph{Proceedings of the IEEE/CVF Conference on Computer
  Vision and Pattern Recognition (CVPR)}}. \bibinfo{pages}{7662--7672}.
\newblock


\bibitem[Thies et~al\mbox{.}(2015)]%
        {thies2015real}
\bibfield{author}{\bibinfo{person}{Justus Thies}, \bibinfo{person}{Michael
  Zollh\"{o}fer}, \bibinfo{person}{Matthias Nie\ss{}ner}, \bibinfo{person}{Levi
  Valgaerts}, \bibinfo{person}{Marc Stamminger}, {and}
  \bibinfo{person}{Christian Theobalt}.} \bibinfo{year}{2015}\natexlab{}.
\newblock \showarticletitle{Real-Time Expression Transfer for Facial
  Reenactment}.
\newblock \bibinfo{journal}{\emph{ACM Trans. Graph.}} \bibinfo{volume}{34},
  \bibinfo{number}{6}, Article \bibinfo{articleno}{183} (\bibinfo{date}{oct}
  \bibinfo{year}{2015}), \bibinfo{numpages}{14}~pages.
\newblock
\showISSN{0730-0301}


\bibitem[Thies et~al\mbox{.}(2016)]%
        {thies2016face2face}
\bibfield{author}{\bibinfo{person}{Justus Thies}, \bibinfo{person}{Michael
  Zollhofer}, \bibinfo{person}{Marc Stamminger}, \bibinfo{person}{Christian
  Theobalt}, {and} \bibinfo{person}{Matthias Niessner}.}
  \bibinfo{year}{2016}\natexlab{}.
\newblock \showarticletitle{Face2Face: Real-Time Face Capture and Reenactment
  of RGB Videos}. In \bibinfo{booktitle}{\emph{Proceedings of the IEEE/CVF
  Conference on Computer Vision and Pattern Recognition (CVPR)}}.
  \bibinfo{pages}{2387--2395}.
\newblock


\bibitem[Vlasic et~al\mbox{.}(2005)]%
        {vlasic2005face}
\bibfield{author}{\bibinfo{person}{Daniel Vlasic}, \bibinfo{person}{Matthew
  Brand}, \bibinfo{person}{Hanspeter Pfister}, {and} \bibinfo{person}{Jovan
  Popovi\'{c}}.} \bibinfo{year}{2005}\natexlab{}.
\newblock \showarticletitle{Face Transfer with Multilinear Models}.
\newblock \bibinfo{journal}{\emph{ACM Trans. Graph.}} \bibinfo{volume}{24},
  \bibinfo{number}{3} (\bibinfo{date}{jul} \bibinfo{year}{2005}),
  \bibinfo{pages}{426–433}.
\newblock
\showISSN{0730-0301}


\bibitem[Wang et~al\mbox{.}(2022)]%
        {wang2022morf}
\bibfield{author}{\bibinfo{person}{Daoye Wang}, \bibinfo{person}{Prashanth
  Chandran}, \bibinfo{person}{Gaspard Zoss}, \bibinfo{person}{Derek Bradley},
  {and} \bibinfo{person}{Paulo Gotardo}.} \bibinfo{year}{2022}\natexlab{}.
\newblock \showarticletitle{MoRF: Morphable Radiance Fields for Multiview
  Neural Head Modeling}. In \bibinfo{booktitle}{\emph{ACM SIGGRAPH 2022
  Conference Proceedings}} (Vancouver, BC, Canada)
  \emph{(\bibinfo{series}{SIGGRAPH '22})}. \bibinfo{publisher}{Association for
  Computing Machinery}, \bibinfo{address}{New York, NY, USA}, Article
  \bibinfo{articleno}{55}, \bibinfo{numpages}{9}~pages.
\newblock
\showISBNx{9781450393379}


\bibitem[Wang et~al\mbox{.}(2020)]%
        {kaisiyuan2020mead}
\bibfield{author}{\bibinfo{person}{Kaisiyuan Wang}, \bibinfo{person}{Qianyi
  Wu}, \bibinfo{person}{Linsen Song}, \bibinfo{person}{Zhuoqian Yang},
  \bibinfo{person}{Wayne Wu}, \bibinfo{person}{Chen Qian}, \bibinfo{person}{Ran
  He}, \bibinfo{person}{Yu Qiao}, {and} \bibinfo{person}{Chen~Change Loy}.}
  \bibinfo{year}{2020}\natexlab{}.
\newblock \showarticletitle{MEAD: A Large-scale Audio-visual Dataset for
  Emotional Talking-face Generation}. In \bibinfo{booktitle}{\emph{Proceedings
  of the European Conference on Computer Vision (ECCV)}}.
\newblock


\bibitem[Wang et~al\mbox{.}(2023)]%
        {wang2023styleavatar}
\bibfield{author}{\bibinfo{person}{Lizhen Wang}, \bibinfo{person}{Xiaochen
  Zhao}, \bibinfo{person}{Jingxiang Sun}, \bibinfo{person}{Yuxiang Zhang},
  \bibinfo{person}{Hongwen Zhang}, \bibinfo{person}{Tao Yu}, {and}
  \bibinfo{person}{Yebin Liu}.} \bibinfo{year}{2023}\natexlab{}.
\newblock \showarticletitle{StyleAvatar: Real-time Photo-realistic Portrait
  Avatar from a Single Video}. In \bibinfo{booktitle}{\emph{ACM SIGGRAPH 2023
  Conference Proceedings}}.
\newblock


\bibitem[Wang et~al\mbox{.}(2021b)]%
        {wang2021one}
\bibfield{author}{\bibinfo{person}{Ting-Chun Wang}, \bibinfo{person}{Arun
  Mallya}, {and} \bibinfo{person}{Ming-Yu Liu}.}
  \bibinfo{year}{2021}\natexlab{b}.
\newblock \showarticletitle{One-Shot Free-View Neural Talking-Head Synthesis
  for Video Conferencing}. In \bibinfo{booktitle}{\emph{Proceedings of the
  IEEE/CVF Conference on Computer Vision and Pattern Recognition (CVPR)}}.
  \bibinfo{pages}{10034--10044}.
\newblock


\bibitem[Wang et~al\mbox{.}(2021a)]%
        {wang2021learning}
\bibfield{author}{\bibinfo{person}{Ziyan Wang}, \bibinfo{person}{Timur
  Bagautdinov}, \bibinfo{person}{Stephen Lombardi}, \bibinfo{person}{Tomas
  Simon}, \bibinfo{person}{Jason Saragih}, \bibinfo{person}{Jessica Hodgins},
  {and} \bibinfo{person}{Michael Zollhofer}.} \bibinfo{year}{2021}\natexlab{a}.
\newblock \showarticletitle{Learning Compositional Radiance Fields of Dynamic
  Human Heads}. In \bibinfo{booktitle}{\emph{Proceedings of the IEEE/CVF
  Conference on Computer Vision and Pattern Recognition (CVPR)}}.
  \bibinfo{pages}{5704--5713}.
\newblock


\bibitem[Weise et~al\mbox{.}(2011)]%
        {weise2011realtime}
\bibfield{author}{\bibinfo{person}{Thibaut Weise}, \bibinfo{person}{Sofien
  Bouaziz}, \bibinfo{person}{Hao Li}, {and} \bibinfo{person}{Mark Pauly}.}
  \bibinfo{year}{2011}\natexlab{}.
\newblock \showarticletitle{Realtime Performance-Based Facial Animation}.
\newblock \bibinfo{journal}{\emph{ACM Trans. Graph.}} \bibinfo{volume}{30},
  \bibinfo{number}{4}, Article \bibinfo{articleno}{77} (\bibinfo{date}{jul}
  \bibinfo{year}{2011}), \bibinfo{numpages}{10}~pages.
\newblock
\showISSN{0730-0301}


\bibitem[Wiles et~al\mbox{.}(2018)]%
        {wiles2018x2face}
\bibfield{author}{\bibinfo{person}{Olivia Wiles}, \bibinfo{person}{A.~Sophia
  Koepke}, {and} \bibinfo{person}{Andrew Zisserman}.}
  \bibinfo{year}{2018}\natexlab{}.
\newblock \showarticletitle{X2Face: A network for controlling face generation
  using images, audio, and pose codes}. In
  \bibinfo{booktitle}{\emph{Proceedings of the Proceedings of the European
  Conference on Computer Vision (ECCV)}}.
\newblock


\bibitem[Xu et~al\mbox{.}(2023)]%
        {xu2023avatarmav}
\bibfield{author}{\bibinfo{person}{Yuelang Xu}, \bibinfo{person}{Lizhen Wang},
  \bibinfo{person}{Xiaochen Zhao}, \bibinfo{person}{Hongwen Zhang}, {and}
  \bibinfo{person}{Yebin Liu}.} \bibinfo{year}{2023}\natexlab{}.
\newblock \showarticletitle{AvatarMAV: Fast 3D Head Avatar Reconstruction Using
  Motion-Aware Neural Voxels}. In \bibinfo{booktitle}{\emph{ACM SIGGRAPH 2023
  Conference Proceedings}}.
\newblock


\bibitem[Yan et~al\mbox{.}(2018)]%
        {yan2018video}
\bibfield{author}{\bibinfo{person}{Shuqi Yan}, \bibinfo{person}{Shaorong He},
  \bibinfo{person}{Xue Lei}, \bibinfo{person}{Guanhua Ye}, {and}
  \bibinfo{person}{Zhifeng Xie}.} \bibinfo{year}{2018}\natexlab{}.
\newblock \showarticletitle{Video Face Swap Based on Autoencoder Generation
  Network}.
\newblock \bibinfo{journal}{\emph{2018 International Conference on Audio,
  Language and Image Processing (ICALIP)}}, \bibinfo{pages}{103--108}.
\newblock


\bibitem[Yariv et~al\mbox{.}(2020)]%
        {yariv2020multiview}
\bibfield{author}{\bibinfo{person}{Lior Yariv}, \bibinfo{person}{Yoni Kasten},
  \bibinfo{person}{Dror Moran}, \bibinfo{person}{Meirav Galun},
  \bibinfo{person}{Matan Atzmon}, \bibinfo{person}{Basri Ronen}, {and}
  \bibinfo{person}{Yaron Lipman}.} \bibinfo{year}{2020}\natexlab{}.
\newblock \showarticletitle{Multiview Neural Surface Reconstruction by
  Disentangling Geometry and Appearance}. In
  \bibinfo{booktitle}{\emph{Conference on Neural Information Processing Systems
  (NeurIPS)}}, Vol.~\bibinfo{volume}{33}.
\newblock


\bibitem[Yenamandra et~al\mbox{.}(2021)]%
        {yenamandra2020i3dmm}
\bibfield{author}{\bibinfo{person}{T Yenamandra}, \bibinfo{person}{A Tewari},
  \bibinfo{person}{F Bernard}, \bibinfo{person}{HP Seidel}, \bibinfo{person}{M
  Elgharib}, \bibinfo{person}{D Cremers}, {and} \bibinfo{person}{C Theobalt}.}
  \bibinfo{year}{2021}\natexlab{}.
\newblock \showarticletitle{i3DMM: Deep Implicit 3D Morphable Model of Human
  Heads}. In \bibinfo{booktitle}{\emph{Proceedings of the IEEE/CVF Conference
  on Computer Vision and Pattern Recognition (CVPR)}}.
\newblock


\bibitem[Yin et~al\mbox{.}(2022)]%
        {yin2022styleheat}
\bibfield{author}{\bibinfo{person}{Fei Yin}, \bibinfo{person}{Yong Zhang},
  \bibinfo{person}{Xiaodong Cun}, \bibinfo{person}{Ming Cao},
  \bibinfo{person}{Yanbo Fan}, \bibinfo{person}{Xuanxia Wang},
  \bibinfo{person}{Qingyan Bai}, \bibinfo{person}{Baoyuan Wu},
  \bibinfo{person}{Jue Wang}, {and} \bibinfo{person}{Yujiu Yang}.}
  \bibinfo{year}{2022}\natexlab{}.
\newblock \showarticletitle{StyleHEAT: One-Shot High-Resolution Editable
  Talking Face Generation via Pre-trained StyleGAN}. In
  \bibinfo{booktitle}{\emph{Proceedings of the European Conference on Computer
  Vision (ECCV)}}.
\newblock


\bibitem[Yu et~al\mbox{.}(2021)]%
        {yu2020pixelnerf}
\bibfield{author}{\bibinfo{person}{Alex Yu}, \bibinfo{person}{Vickie Ye},
  \bibinfo{person}{Matthew Tancik}, {and} \bibinfo{person}{Angjoo Kanazawa}.}
  \bibinfo{year}{2021}\natexlab{}.
\newblock \showarticletitle{pixelNeRF: Neural Radiance Fields from One or Few
  Images}. In \bibinfo{booktitle}{\emph{2021 IEEE/CVF Conference on Computer
  Vision and Pattern Recognition (CVPR)}}. \bibinfo{pages}{4576--4585}.
\newblock


\bibitem[Zakharov et~al\mbox{.}(2019)]%
        {zakharov2019few}
\bibfield{author}{\bibinfo{person}{Egor Zakharov}, \bibinfo{person}{Aliaksandra
  Shysheya}, \bibinfo{person}{Egor Burkov}, {and} \bibinfo{person}{Victor
  Lempitsky}.} \bibinfo{year}{2019}\natexlab{}.
\newblock \showarticletitle{Few-Shot Adversarial Learning of Realistic Neural
  Talking Head Models}. In \bibinfo{booktitle}{\emph{Proceedings of the
  IEEE/CVF International Conference on Computer Vision (ICCV)}}.
  \bibinfo{pages}{9458--9467}.
\newblock


\bibitem[Zhang et~al\mbox{.}(2022)]%
        {zhang2022fdnerf}
\bibfield{author}{\bibinfo{person}{Jingbo Zhang}, \bibinfo{person}{Xiaoyu Li},
  \bibinfo{person}{Ziyu Wan}, \bibinfo{person}{Can Wang}, {and}
  \bibinfo{person}{Jing Liao}.} \bibinfo{year}{2022}\natexlab{}.
\newblock \showarticletitle{FDNeRF: Few-Shot Dynamic Neural Radiance Fields for
  Face Reconstruction and Expression Editing}. In
  \bibinfo{booktitle}{\emph{SIGGRAPH Asia 2022 Conference Papers}} (Daegu,
  Republic of Korea) \emph{(\bibinfo{series}{SA '22})}.
  \bibinfo{publisher}{Association for Computing Machinery},
  \bibinfo{address}{New York, NY, USA}, Article \bibinfo{articleno}{12},
  \bibinfo{numpages}{9}~pages.
\newblock
\showISBNx{9781450394703}


\bibitem[Zhang et~al\mbox{.}(2016)]%
        {yucheng2019joint}
\bibfield{author}{\bibinfo{person}{Kaipeng Zhang}, \bibinfo{person}{Zhanpeng
  Zhang}, \bibinfo{person}{Zhifeng Li}, {and} \bibinfo{person}{Yu Qiao}.}
  \bibinfo{year}{2016}\natexlab{}.
\newblock \showarticletitle{Joint Face Detection and Alignment Using Multitask
  Cascaded Convolutional Networks}.
\newblock \bibinfo{journal}{\emph{IEEE Signal Processing Letters}}
  \bibinfo{volume}{23}, \bibinfo{number}{10} (\bibinfo{year}{2016}),
  \bibinfo{pages}{1499--1503}.
\newblock


\bibitem[Zhang et~al\mbox{.}(2018)]%
        {zhang2018the}
\bibfield{author}{\bibinfo{person}{Richard Zhang}, \bibinfo{person}{Phillip
  Isola}, \bibinfo{person}{Alexei~A. Efros}, \bibinfo{person}{Eli Shechtman},
  {and} \bibinfo{person}{Oliver Wang}.} \bibinfo{year}{2018}\natexlab{}.
\newblock \showarticletitle{The Unreasonable Effectiveness of Deep Features as
  a Perceptual Metric}. In \bibinfo{booktitle}{\emph{Proceedings of the
  IEEE/CVF Conference on Computer Vision and Pattern Recognition (CVPR)}}.
  \bibinfo{pages}{586--595}.
\newblock


\bibitem[Zheng et~al\mbox{.}(2022)]%
        {zheng2022imavatar}
\bibfield{author}{\bibinfo{person}{Yufeng Zheng},
  \bibinfo{person}{Victoria~Fern\'andez Abrevaya}, \bibinfo{person}{Marcel~C.
  B\"uhler}, \bibinfo{person}{Xu Chen}, \bibinfo{person}{Michael~J. Black},
  {and} \bibinfo{person}{Otmar Hilliges}.} \bibinfo{year}{2022}\natexlab{}.
\newblock \showarticletitle{I M Avatar: Implicit Morphable Head Avatars From
  Videos}. In \bibinfo{booktitle}{\emph{Proceedings of the IEEE/CVF Conference
  on Computer Vision and Pattern Recognition (CVPR)}}.
  \bibinfo{pages}{13535--13545}.
\newblock


\bibitem[Zheng et~al\mbox{.}(2023a)]%
        {zheng2023pointavatar}
\bibfield{author}{\bibinfo{person}{Yufeng Zheng}, \bibinfo{person}{Wang Yifan},
  \bibinfo{person}{Gordon Wetzstein}, \bibinfo{person}{Michael~J. Black}, {and}
  \bibinfo{person}{Otmar Hilliges}.} \bibinfo{year}{2023}\natexlab{a}.
\newblock \showarticletitle{PointAvatar: Deformable Point-based Head Avatars
  from Videos}. In \bibinfo{booktitle}{\emph{Proceedings of the IEEE/CVF
  Conference on Computer Vision and Pattern Recognition (CVPR)}}.
\newblock


\bibitem[Zheng et~al\mbox{.}(2023b)]%
        {zheng2023avatarrex}
\bibfield{author}{\bibinfo{person}{Zerong Zheng}, \bibinfo{person}{Xiaochen
  Zhao}, \bibinfo{person}{Hongwen Zhang}, \bibinfo{person}{Boning Liu}, {and}
  \bibinfo{person}{Yebin Liu}.} \bibinfo{year}{2023}\natexlab{b}.
\newblock \showarticletitle{AvatarReX: Real-time Expressive Full-body Avatars}.
\newblock \bibinfo{journal}{\emph{ACM Transactions on Graphics (TOG)}}
  \bibinfo{volume}{42}, \bibinfo{number}{4} (\bibinfo{year}{2023}),
  \bibinfo{pages}{1--19}.
\newblock
\urldef\tempurl%
\url{https://doi.org/10.1145/3592101}
\showDOI{\tempurl}


\bibitem[Zhu et~al\mbox{.}(2017)]%
        {zhu2017unpaired}
\bibfield{author}{\bibinfo{person}{Jun-Yan Zhu}, \bibinfo{person}{Taesung
  Park}, \bibinfo{person}{Phillip Isola}, {and} \bibinfo{person}{Alexei~A.
  Efros}.} \bibinfo{year}{2017}\natexlab{}.
\newblock \showarticletitle{Unpaired Image-to-Image Translation Using
  Cycle-Consistent Adversarial Networks}.
\newblock \bibinfo{journal}{\emph{Proceedings of the IEEE/CVF International
  Conference on Computer Vision (ICCV)}}, \bibinfo{pages}{2242--2251}.
\newblock


\bibitem[Zhuang et~al\mbox{.}(2022)]%
        {zhuang2022mofanerf}
\bibfield{author}{\bibinfo{person}{Yiyu Zhuang}, \bibinfo{person}{Hao Zhu},
  \bibinfo{person}{Xusen Sun}, {and} \bibinfo{person}{Xun Cao}.}
  \bibinfo{year}{2022}\natexlab{}.
\newblock \showarticletitle{MoFaNeRF: Morphable Facial Neural Radiance Field}.
  In \bibinfo{booktitle}{\emph{Proceedings of the European Conference on
  Computer Vision (ECCV)}}.
\newblock


\bibitem[Zielonka et~al\mbox{.}(2022)]%
        {zielonka2022instant}
\bibfield{author}{\bibinfo{person}{Wojciech Zielonka}, \bibinfo{person}{Timo
  Bolkart}, {and} \bibinfo{person}{Justus Thies}.}
  \bibinfo{year}{2022}\natexlab{}.
\newblock \bibinfo{title}{Instant Volumetric Head Avatars}.
\newblock
\newblock
\showeprint[arxiv]{2211.12499}~[cs.CV]


\end{thebibliography}


\end{document}